%% file: acl_latex.tex
\pdfoutput=1
\documentclass[11pt]{article}

\usepackage[preprint]{acl}

\usepackage{times}
\usepackage{latexsym}
\usepackage{booktabs}

\usepackage[T1]{fontenc}
\usepackage[utf8]{inputenc}

\usepackage{microtype}

\usepackage{inconsolata}

\usepackage{graphicx}
\usepackage{amsmath}
\usepackage{amssymb}
\usepackage{caption}
\usepackage{amsthm}
\usepackage{enumitem}
\usepackage{lipsum}
\usepackage{bm}
\usepackage{newpxmath}
\usepackage{mathmacros}

\title{Circuit Stability Characterizes Language Model Generalization}

\author{Alan Sun\\
  Carnegie Mellon University \\
  \texttt{alansun@andrew.cmu.edu} 
}

\newcommand\opra{\text{Opr}_1}
\newcommand\oprb{\text{Opr}_2}

\newtheorem{defn}{Definition}

\begin{document}
\maketitle

\begin{abstract}
Extensively evaluating the capabilities of (large) language models is difficult. Rapid development of state-of-the-art models induce benchmark saturation, while creating more challenging datasets is labor-intensive. Inspired by the recent developments in mechanistic interpretability, we introduce \textit{circuit stability} as a new way to assess model performance. Circuit stability refers to a model's ability to apply a consistent reasoning process---its circuit---across various inputs. We mathematically formalize circuit stability and \textit{circuit equivalence}. Then, through three case studies, we empirically show that circuit stability and the lack thereof can characterize and predict different aspects of generalization. Our proposed methods offer a step towards rigorously relating the generality of models to their interpretability\footnote{The codebase for our experiments can be found at \url{https://github.com/alansun17904/circuit-stability}}. 

\end{abstract}

\input{sections/intro}

\input{sections/prelims}

\input{sections/circuit-stability}

\input{sections/arithmetic}

\input{sections/boolean}

\input{sections/prompting}

\input{sections/related}

\input{sections/discussion}

\input{appendix/limitations}

\section*{Acknowledgments}
Alan thanks Mariya Toneva for her generous support; Fengwen Sun for his valuable feedback and advice; Ethan Sun and Warren Shepard for their contributions to early iterations of the work; and, Andrew Koulogeorge for the thoughtful discussions.

\vspace{-0.4cm}

\bibliography{references}

\appendix

\input{appendix/beyond}
% \input{appendix/technical}
\input{appendix/patching}
\input{appendix/napkin}

\input{appendix/reproduction}

\end{document}

%% file: sections/intro.tex
% !TEX root = ../acl_latex.tex

\section{Introduction}

Though there exists a wealth of theoretical techniques to analyze and predict generalization, empirically benchmarking a given language model's capabilities remains difficult. This gap between theory and practice stems, in part, from the rapid saturation of existing benchmarks and also the labor-intensive process of creating new, more challenging datasets~\cite{srivastava2023beyond,jimenezswe,glazer2024frontiermath}. One way to sidestep these issues is to evaluate a specific capability shared among many tasks. For example, needle-in-the-haystack evaluates language models' ability to perform long-context recall~\cite{kamradt2023needle}, while \textsc{skill-mix} measures skill composition performance~\cite{arora2023theory,yu2024skillmix}. Although such datasets can be automatically generated and scaled in difficulty, a fundamental challenge remains: identifying specific capabilities that are meaningful to benchmark in the first place. Even after identifying a capability-of-interest, creating salient tasks that precisely target this capability is nontrivial. 

In this paper, we seek to address some of these issues by introducing the concept of \textit{circuit stability}. Informally, for a fixed task and model, circuit stability is the consistency of a model's reasoning process (its circuit) across collections of subtasks. Consider solving an algorithmic problem like parity. We expect a strong model to learn and correctly apply a consistent algorithm regardless of the length of its input. If, however, the model learns a different algorithm for each input length, its finite capacity will guarantee a length past which the model will fail. 

At the core of our approach are three key insights. 
First, orthogonal to previous approaches that evaluate a model's performance on small, finite sets of examples by testing inputs one-by-one, we extract and analyze the model's circuit using techniques from mechanistic interpretability~\cite{olah2020zoom}. Since the extracted circuit can be reused and applied by the model to an infinite class of examples, its stability could be a more robust estimator of performance. This is analogous to the formal verification of an algorithm where we do not need to verify every input-output pair~\cite{cousot1977abstract}. 
Additionally, instead of specifying individual, potentially contentious skills-of-interest, our approach makes a simplifying assumption that a learned skill/circuit is useful only if it is consistently applied by the model. 
Finally, unlike mechanistic interpretability approaches that seek to intractably extract \textit{hard circuits}, which are discrete subsets of the model's computational graph, we introduce a continuous relaxation: \textit{soft circuits}. This preserves rich structural insights while enabling more easy computation. 

Concretely, our contributions are four-fold:
\begin{itemize}[nosep, leftmargin=0.05\linewidth]
\item We formally define circuit stability (Section~\ref{sec:circuit_stability}). 
\item Through two case studies on arithmetic reasoning and Boolean expression evaluation, we show that circuit stability can predict length, structural, and compositional generalization (Section~\ref{sec:case_study_arithmetic_reasoning} and Section~\ref{sec:case_study_boolean_expression_evaluation}, respectively).
\item We show that circuit stability also predicts generalization even on tasks that are not naturally algorithmic like sports understanding~\cite{suzgun-etal-2023-challenging}.
\item We demonstrate that circuit stability can be induced through prompting methods like chain-of-thought (Section~\ref{sec:case_study_prompting_effectiveness}).
\end{itemize}
In this paper, we focus our analyses of circuit stability on language models based on the Transformer architecture~\cite{vaswani_attention_2017}. Nevertheless, the formal framework we introduce is modality- and architecture-independent. We discuss this further in Appendix~\ref{app:beyond}.

%% file: sections/prelims.tex
% !TEX root = ../acl_latex.tex

\section{Background}\label{sec:background}
Herein, we briefly review the circuits framework and relevant concepts in mechanistic interpretability, as they lay the foundation for our contribution. 

\noindent\textbf{Circuits.} The goal of the circuits framework is to interpret the decision processes of a neural network. This is typically done through two processes: first, identify a minimal subset of the model's computational graph that is responsible for a specific behavior; second, assign human-interpretable explanations to each of the extracted components. The former is referred to as \textit{circuit discovery} while the latter is called \textit{mechanistic interpretability}. 

We represent a Transformer's computational graph using the framework introduced by~\cite{elhage_mathematical_2021, conmy_towards_2023}. Specifically, we view each MLP layer as single node and, unless otherwise specified, we also split each attention head into four distinct nodes: key, query, value, and output. A directed edge is drawn from nodes $n_i \to n_j$ if the output of $n_i$ is directly used as an input to $n_j$. A circuit is defined as a computational subgraph. 

\noindent\textbf{Finding Subcircuits.} Let $G^M = (V^M,E^M)$ be the computational graph for a model $M$. For a fixed task and model performance metric $L$, circuit discovery methods search for a \textit{minimal} computational subgraph: $g^M = (v^M, e^M)$, where $v^M \subset V^M, e^M \subset E^M$. Informally, after ablating the edges and nodes not in $g^M$, $g^M$ must maintain model performance within a specified margin of $\eps > 0$~\cite{wang_interpretability_2022,shi2024hypothesis}.
Alternatively, one can also view this process as searching for a binary function, $c: E^M \to \{0,1\}$, subject to the aforementioned constraints. We refer to $c$ as a \textit{hard circuit} because any given edge in the computational graph takes on a binary state. Searching for $c$ is known to be an intractable combinatorial optimization problem~\cite{adolfi2025the}. As a result, many approximations to circuit discovery have been derived~\cite{nanda_attribution_2023,hanna_have_2024,bhaskar_finding_2024}. Even then, it has been shown that the discovered circuit may be highly sensitive to $L$ and $\eps$, the performance metric and threshold, respectively~\cite{conmy_towards_2023,miller2024transformer}. In this paper, we partially circumvent these issues by redefining a circuit as a mapping $c: E^M \to \R$. For each $e \in E^M$, $c(e)$ represents the change in $L$ after ablating $e$ from $E^M$, essentially capturing the importance of $e$. In this way, we yield a \textit{soft circuit}. We formalize this in Section~\ref{sec:circuit_stability}.

%% file: sections/circuit-stability.tex
% !TEX root = ../acl_latex.tex

\section{Circuit Stability and Equivalence}\label{sec:circuit_stability}

% definition 1,3, 

In this section, we formally define circuit stability. First, we define the notion of a task (Definition~\ref{def:task}). Then, tasks are equipped with variable substructure through subtasks (Definition~\ref{def:subtask}). Building on these subtasks, we define the three key concepts of this paper: soft circuitry (Definition~\ref{def:circuit}), $\eps$-circuit stability (Definition~\ref{def:circuit-stable}) and $\alpha$-circuit equivalence (Definition~\ref{def:circuit-equivalence}).

Let $\cX, \cY$ be the space of all finite input and output strings~\cite{du-etal-2023-measure,cotterell2023formal}. The exact construction of these spaces are unimportant. We only require a distribution over $\cX \times \cY$ which we call a task.  
\begin{defn}[Task]\label{def:task}
A \textbf{task} is a distribution over $\cX \times \cY$ denoted by $\data$. This is also called the \textbf{data distribution}.
\end{defn}
A task may itself contain rich substructure. For even a simple task such as two-operand addition\footnote{In this case, the marginal of $\data$ over $\cX$ assigns positive measure to a subset $X \subset \cX$ only if all $x \in X$ follows the form \texttt{``a + b = ''} where $a, b \in \Z^{\geq 0}$. And, the conditional distribution of $\data$ on $\cY$ given \texttt{``a + b = ''} is a point mass on the string $a + b$.}, we can, naturally decompose this into many distinct collections of subtasks by simply partitioning the input-output space. One way is to separate addition problems that require carrying at least one digit versus ones that do not. On the other hand, we could also create another collection of subtasks by varying the number of digits in each operand. Intuitively, an appropriate partitioning of the input-output space should yield a collection of subtasks that make clear the necessary capabilities a model must have to solve the task itself. For example, success over the aforementioned partitions could indicate both compositional and length generalization, respectively~\cite{wiedemer2023compositional}. Without the loss of generality, we assume that all cells of any subsequently discussed partition have positive measure under $\data$. In this way, each cell contains meaningful examples and skills that a model must master in order to achieve perfect performance. By way of its cells, a given partition also elicits its own set of tasks. We call these conditional distributions \textit{subtasks}. 
\begin{defn}[Subtask]\label{def:subtask}
For a task, $\data$ and partition $\cS$ of $\cX \times \cY$. A subtask, over a cell $s \in \cS$,  is the conditional distribution $\data | s$. For brevity, we notate this as $\cD_s$.
\end{defn}

Through $\data$, we can also measure the importance of any subtask by leveraging the marginal distribution over $\cS$. We call this distribution the \textit{partition distribution} and notate it as $\P_\cS$.

We now precisely define a model's \textit{soft circuit} relative to a task (or subtask). Our approach differs from the traditional notion of a circuit defined in mechanistic interpretability. Rather than assign a binary indicator to the edges of the computational graph, $E^M \to \{0,1\}$, we perform a continuous relaxation. A comparative analysis of this setting, along with its implications, are provided in Section~\ref{sec:discussion}.
% we find the minimal circuit Occam's razor
\begin{defn}[Soft Circuit]\label{def:circuit}
Consider a task $\data$, a model $M: \cX \to \cY$ with computational graph $G^M = (V^M,E^M)$, and some performance metric $L: \cY \times \cY \to \R$. With respect to $\data$, $M$'s \textbf{soft circuit} is a function $c: E^M\to \R$ such that for any $e \in E^M$, 
\begin{equation}\label{eq:comp-intervention}
c(e) \coloneqq \mspace{-25mu}\underset{{\mspace{30mu}(x,y) \sim \data}}{\E}\mspace{-25mu}[L(M_{\{e\}}(x) , y)-L(M(x), y)],
\end{equation}
where $M_{\{e\}}$ denotes $M$ after ablating edge $e$. 
\end{defn}
As long as $L$ is well-defined, $c$ always exists. We defer the technical details of $L$ and the ablation procedure to Appendix~\ref{app:circuit_discovery_details}. By our previous constructions, a subtask (Definition~\ref{def:subtask}) may also induce a soft circuit. So for any subtask $\cD_s$, we denote its induced soft circuit as $c_s$.

Intuitively, $c(e)$ captures the singular importance of $e$. By examining and comparing the collective mapping, $c$, across subtasks, we gain insight into holistic model behavior. Therefore, we take $K: \R^{E^M} \times \R^{E^M} \to \R$ to be a measure of the similarity between two soft circuits: a kernel-like function\footnote{$K$ should be thought of as a reproducing kernel Hilbert space kernel over the function space $\R^{E_M}$. However, for simplicity, our experiments do not adhere to this guiding principle. Instead we use a more interpretable similarity metric such as rank correlation. A deeper investigation into the theoretical properties of $K$ and its implications for circuit stability is left for future work}. We are now ready to define circuit stability.

\begin{defn}[$\eps$-Circuit Stable]\label{def:circuit-stable}
For $\eps > 0$, a model, $M$, is \textbf{$\bm{\eps}$-circuit stable} with respect to a task $\data$ and a collection of partitions $\sP$ if
\begin{equation}\label{eq:circuit-stable}
\inf_{\cS \in \sP} \E_{s, s' \sim \cS} [K(c_s, c_{s'})] > \eps,
\end{equation}
where $s,s'$ are two subtasks sampled i.i.d. from the partition distribution
$\P_\cS$. 
\end{defn}
In Equation~\ref{eq:circuit-stable}, the expression inside the infimum measures the stability of a model's soft circuit as we move between different subtasks. This stability is weighted by the partition distribution. Thus, if two subtasks have a low probability of occurring with respect to $\data$, we consider the skills required to solve them unimportant. In turn, instability across these subtasks is also disregarded. This allows us to avoid specifying \textit{a priori} which skills are important for analysis. 

For all our experiments, we take $K$ to be Spearman's $\rho$. Concretely, we take soft circuits $c_s, c_{s'}$ and individually induce a ranking of $E^M$ through $c_s(E^M), c_{s'}(E^M)$. Then, we measure the correlation coefficient between these ranks. Throughout, we construct collections of partitions $\sP$ manually, based on the task at hand. In Section~\ref{sec:case_study_prompting_effectiveness} and~\ref{sec:discussion}, we discuss constructing partitions statistically.

Next, using $K$ we also define a type of pointwise-equivalence between the soft circuits of any two subtasks. 
\begin{defn}[$\alpha$-Equivalent]\label{def:circuit-equivalence}
For $\alpha > 0$, soft circuits
$c_s, c_{s'}$ are \textbf{$\alpha$-equivalent} if $K(c_s, c_s') \geq \alpha$.  
\end{defn}

%% file: sections/arithmetic.tex
% !TEX root = ../acl_latex.tex

\section{Case Study: Arithmetic Reasoning}\label{sec:case_study_arithmetic_reasoning}

In this section, we explore the circuit stability and equivalence of \texttt{gemma-2-2b}\footnote{2 billion parameter model containing over 79k circuit edges as defined in Section~\ref{sec:background}.} over the task of two operand addition~\cite{gemma}. These problems come in the form of \texttt{``a + b = ''}, where $a, b \in \Z^{\geq 0}$. We examine a specific partition where each subtask contains problems where the values of $a$ all have the same number of digits, and similarly for $b$. This setup allows us to study two different forms of generalization that have been separately analyzed in the literature: length generalization, where the number of digits in $a,b$ increase independently~\cite{cho2025arithmetic}, and compositional generalization which tests whether models solve addition problems recursively~\cite{kudo-etal-2023-deep,nikankin2025arithmetic}. We find that \texttt{gemma-2-2b}'s circuit instability across subtasks correlates with fluctuations in its performance on those same subtasks, indicating a potential causal link between circuit instability and generalization failures.

\subsection{Experimental Setup}\label{sec:arithmetic-experimental-setup}
We denote a subtask as an ordered pair $(o_1, o_2)$ where $1 \leq o_1, o_2 \leq 8$. 
$o_1, o_2$ denote the number of digits in $a,b$ respectively.
Over all experimental settings, we provide the model with $k=3$ few-shot examples. To implement circuit discovery (Definition~\ref{def:circuit}), we choose $L$ to be the next-token patching metric, defined in Equation~\ref{eq:next-token-patching}. Each edge is ablated through noisy-to-clean patching using both noisy and clean samples from the same subtask. These design choices are well-documented in~\citet{heimersheim_how_2024} and we explore their implications in Appendix~\ref{app:circuit_discovery_details}. We perform circuit discovery over each subtask, resulting in 64 soft circuits for analysis.

\subsection{Identifying Arithmetic Circuit Families}

\begin{figure}
\centering
\includegraphics[width=0.65\linewidth]{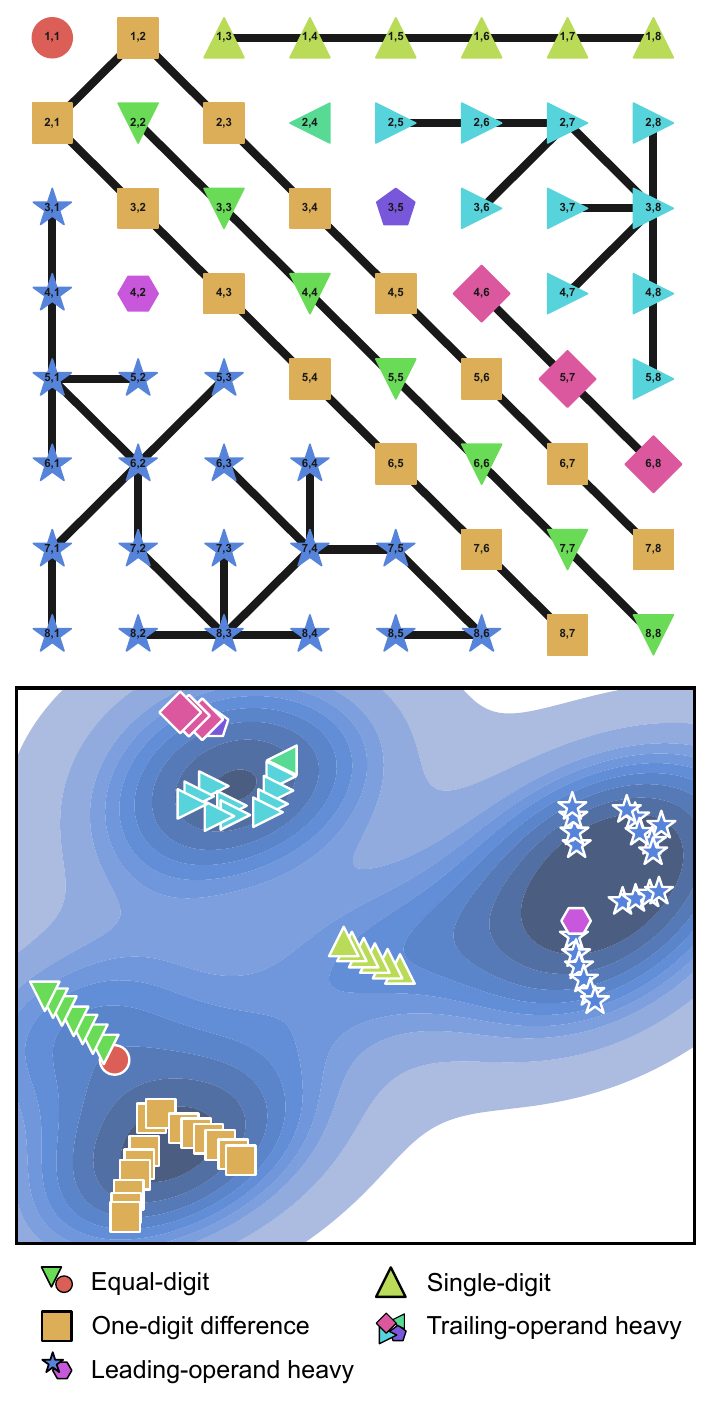}
\caption{\textbf{(top)} Each node represents a distinct subtask. An undirected path exists between two nodes if and only if they are $\alpha$-equivalence for $\alpha=0.6$.
\textbf{(bot)} $t$-SNE embeddings of the soft circuits for each subtask with \texttt{perplexity=3}. The node shape and color combinations are consistent with the circuit clusters in (top).}
\label{fig:tsned-structures}
\vspace{-0.6cm}
\end{figure}

To analyze the relationships between subtasks, we compute Spearman's $\rho$, $K$, between their soft circuits. Using $\alpha=0.6$, we apply the notion of $\alpha$-equivalence, as defined in Definition~\ref{def:circuit-equivalence}, to cluster the arithmetic subtasks into roughly five distinct clusters: equal-digit ($o_1 = o_2$), one-digit difference ($o_1 = o_2 \pm 1$), leading-operand heavy ($o_1 > o_2$), single-digit ($o_2 = 1$), and trailing-operand heavy ($o_2 > o_1$). These subtask clusters are visualized in Figure~\ref{fig:tsned-structures}\textbf{(top)}. 

%% TODO: as defined in section, ...

To confirm that these clusters are not merely an artifact of a particular setting of $\alpha$, we perform two complementary experiments. First, we directly compute a set of $t$-SNE embeddings using all the soft circuits~\cite{maaten_visualizing_2008}. Notably, these embeddings are independent of $\alpha$. The relative distances between the embedded circuits are visualized in Figure~\ref{fig:tsned-structures}\textbf{(bot)}. We find that the circuit clusters we identified before form well-separated groups under this representation as well. 

\begin{figure}[h]
\centering
\includegraphics[width=0.8\linewidth]{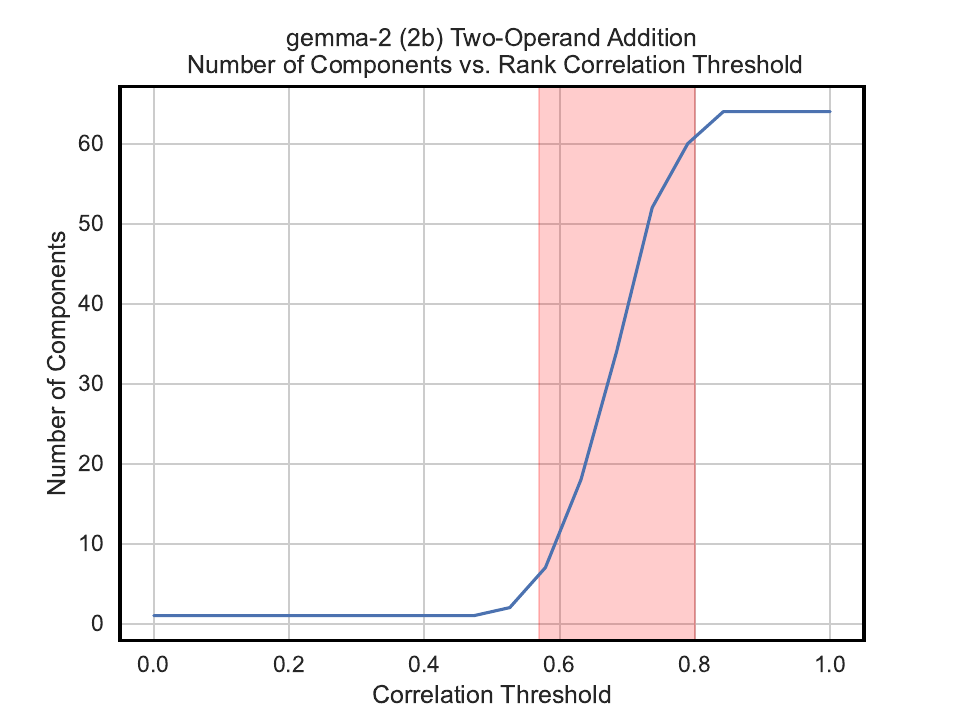}
\caption{The number of $\alpha$-circuit equivalence families as $\alpha$ varies between $[0,1]$. We omit visualization of $\alpha < 0$ since the number of circuit families is a monotonic function of $\alpha$. The red region shows that 80\% of circuit families emerge between $\alpha=0.58$ and $\alpha=0.79$}
\label{fig:families-v-alpha}
\vspace{-0.2cm}
\end{figure}

Next, since we take $K$ to be Spearman's $\rho$, $\alpha$ is naturally bounded between $[-1,1]$. Consequently, we expect the number of distinct subtask clusters to increase monotonically as $\alpha$ also increases monotonically. This behavior is shown in Figure~\ref{fig:families-v-alpha}. As $\alpha$ approaches its upper bound of 1, the number of subtask clusters converges to the total number of subtasks. This indicates that no two subtasks have identical circuits. On the other hand, for $\alpha = 0.4$, only one subtask cluster exists. This suggests the presence of a common set of circuit components that are important for arithmetic generally. In other words, no two distinct subtasks reply on entirely disjoint sets of components. This observation aligns with the previous findings of~\citet{stolfo-etal-2023-mechanistic,hanna_how_2023,nikankin2025arithmetic} that the numerical abilities of language models are mediated by a common set of attention heads and MLPs.

\begin{figure}
\centering
\includegraphics[width=0.8\linewidth]{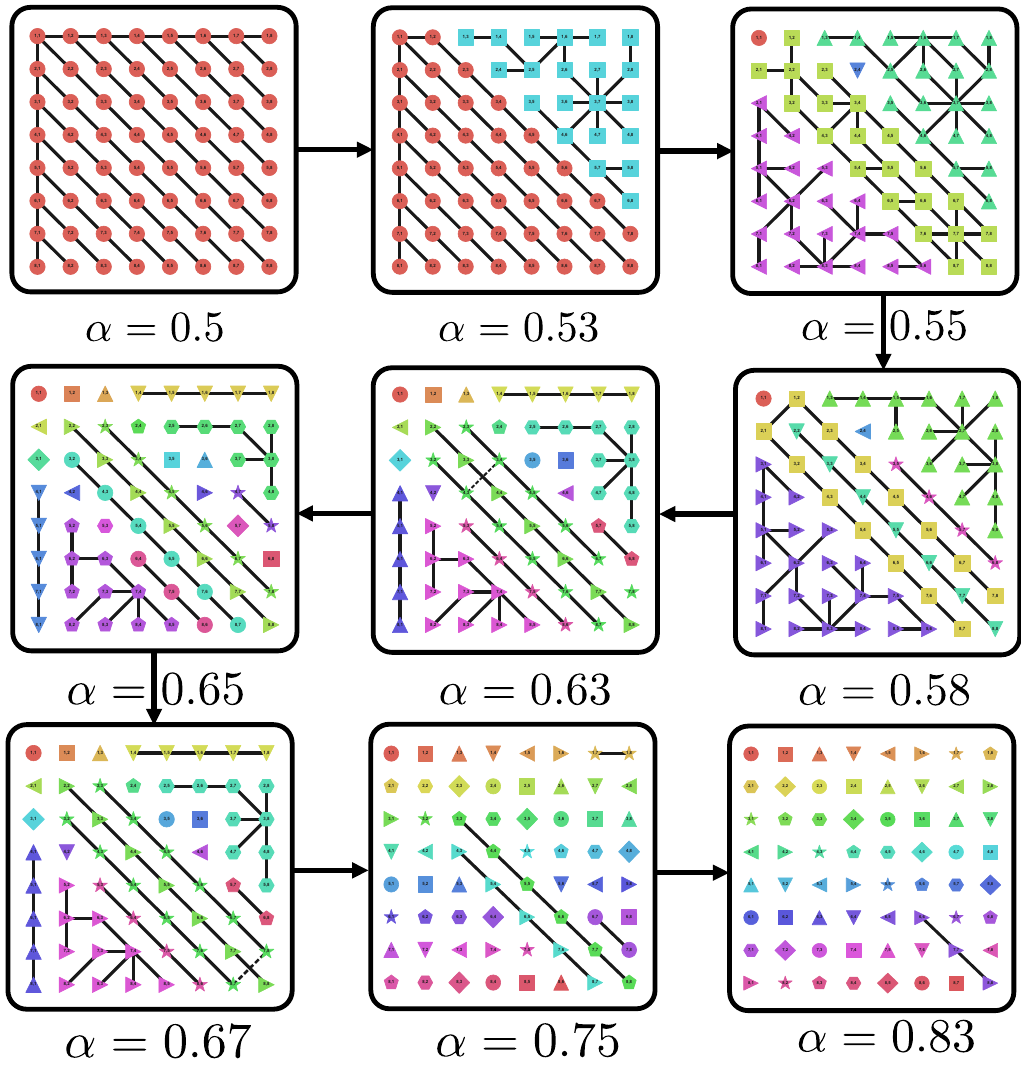}
\caption{The emergent families of $\alpha$-equivalent subtasks as $\alpha$ varies between $[0,1]$.}
\label{fig:subtask-family-a}
\vspace{-0.5cm}
\end{figure}

Surprisingly, there exists a critical threshold $\alpha = 0.6$ where the number of circuit families explodes (see the red region in Figure~\ref{fig:families-v-alpha}). We visualize the circuit families in this critical region in Figure~\ref{fig:subtask-family-a}. As $\alpha$ increases from 0.5 to 0.53, a clear separation immediately forms between the trailing-operand and leading-operand heavy subtasks. This could indicate fundamental differences in how the model is handling inputs from these two subtasks. As $\alpha$ continues to increase, we observe the emergence and persistence of distinct families corresponding to equal-digit, single-digit, and one-digit difference subtasks. The stability of these families over increasing $\alpha$ suggests strong internal cohesion, further validating our proposed clustering depicted in Figure~\ref{fig:tsned-structures}\textbf{(bot)}.

\subsection{Circuit Stability and Generalization}
The family of $\alpha$-equivalences (Definition~\ref{def:circuit-equivalence}) identified in the previous subsection deviate from our expected clustering of a well-performing model (Section~\ref{sec:circuit_stability}). In particular, a model that truly understands two-operand addition should necessarily recognize that addition is both commutative and associative. In this section, we argue that lack of $\alpha$-equivalence across these aforementioned circuit families indicates that the model neither adheres to nor fully internalizes these properties of addition. 

\begin{figure}
\centering
\includegraphics[width=0.8\linewidth]{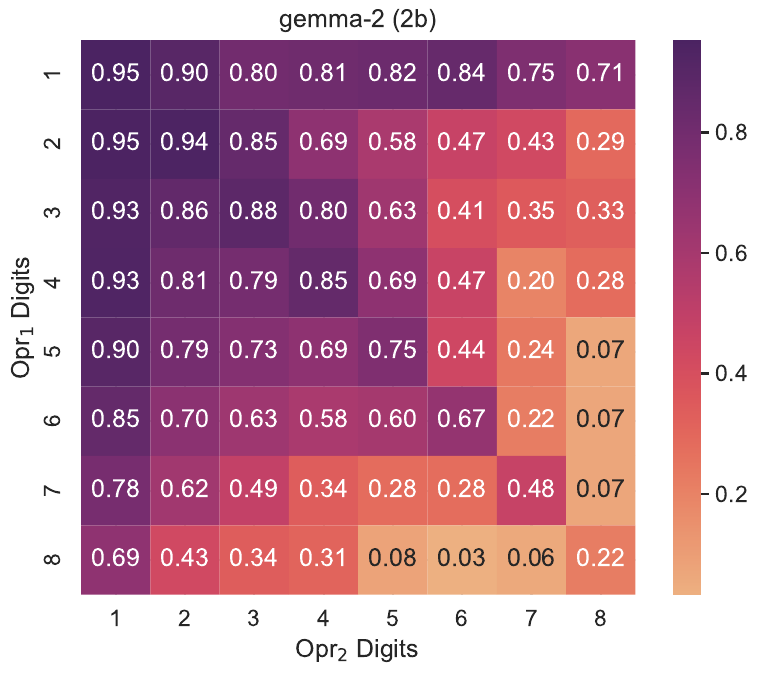}
\vspace{-0.1cm}
\caption{Performance of \texttt{gemma-2-2b} over arithmetic subtasks as $\opra \coloneqq o_1$ and $\oprb \coloneqq o_2$ increase. Each cell denotes the exact string match accuracy.}
\label{fig:benchmark}
\vspace{-0.5cm}
\end{figure}

First, an understanding and application of commutativity implies that any subtask $(o_1, o_2)$ should be equivalent both in performance and circuitry to the subtask $(o_2, o_1)$.
This is because a model that learns this axiom could accordingly transpose the two operands before adding, thereby achieving consistent performance across this collections of subtasks. However, as discussed previously, a lack of $\alpha$-equivalence between leading and trailing-digit heavy families suggest that \texttt{gemma-2-2b} does not respect commutativity. As a result, we expect a significant performance gap as we move between these subtask families. Indeed we observe this to be the case. We benchmark \texttt{gemma-2-2b} across all 64 subtasks. Per task, $n=1000$ problems are sampled independently while maintaining the same formatting scheme as before (see Section~\ref{sec:arithmetic-experimental-setup}). Performance is measured through exact string match accuracy~\cite{srivastava2023beyond}. We find that \texttt{gemma-2-2b}'s performance positively skews towards leading-digit heavy subtasks (see Figure~\ref{fig:benchmark}). In some cases, the performance difference between $(o_1, o_2)$ and $(o_2, o_1)$ can be more than 20\%.

Next, the associativity of addition implies that any two-operand addition problem can be decomposed into a sequence of $(1,1)$ problems. More generally, a subtask like $(8,2)$ could also be broken down into a sequence of $(8,1)$ and $(1,1)$ problems. Likewise, $(6,7)$ can be decomposed into $(6,6)$ and $(1,1)$. If the model is leveraging associativity, we would expect its errors to also compound in a predictable manner due to the repeated reuse of simpler subtasks. But, we observe through the previous subsection, that even adjacent subtasks like $(8,1)$ vs. $(8,2)$ or $(6,6)$ vs. $(6,7)$ belong to different $\alpha$- equivalent clusters. This suggests that \texttt{gemma-2-2b} is not exploiting associativity to systemically reuse its circuit components across subtasks. This behavior is verified quantitatively in Figure~\ref{fig:benchmark}, where model performance steeply drops off across subtasks $(6,6), (6,7)$, etc.

\begin{figure}
\centering
\includegraphics[width=0.9\linewidth]{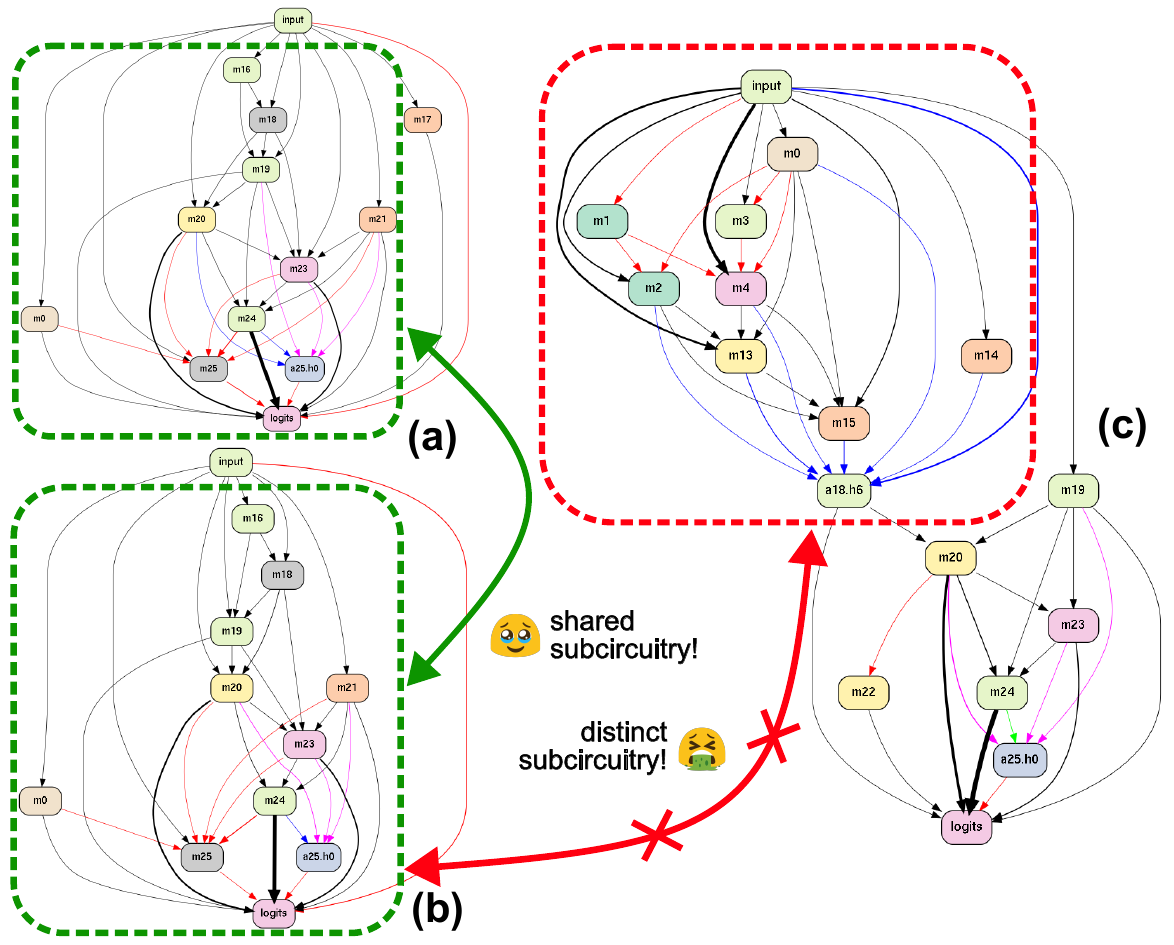}
\caption{Circuits for \textbf{(a)} $(8,8)$ \textbf{(b)} $(7,7)$ and \textbf{(c)} $(2,7)$. \textbf{(a)} and \textbf{(b)} share many subcircuit components and are $\alpha$-equivalent for $\alpha = 0.6$. On the other hand, (c) is not $\alpha$-circuit equivalent with either (a) or (b).}
\label{fig:shared-subcircuitry}
\vspace{-0.6cm}
\end{figure}

Lastly, we hypothesize that within an $\alpha$-equivalent cluster, \texttt{gemma-2-2b} \textit{is} reusing its circuit components. This behavior has been previously identified in other tasks~\cite{merullo_circuit_2023}. We find that hard circuits within the same $\alpha$-equivalent cluster share a large number of components or, in some cases, even function as subcircuits of one another (see an example in Figure~\ref{fig:shared-subcircuitry}). Here, we greedily construct\footnote{Though this is a naive decoding method, empirically it works quite well as an approximation for the actual circuit~\cite{conmy_towards_2023,hanna_have_2024}. } hard circuits by assigning the top 200 components---as given by the soft circuitry---one (in circuit) and the remaining zero (out of circuit). In contrast to the sharp performance change across non-equivalent subtasks, we find that within $\alpha$-equivalent subtasks model performance decays smoothly. This degradation in performance can be characterized using tight-fitting regressions that depend on the number of subtask compositions and their associated error rates. We provide a detailed analysis of this in Appendix~\ref{app:napkin_calcuations}. 

By combining our insights derived from circuit stability analysis with the benchmark results in Figure~\ref{fig:benchmark}, we can affirm that the measured performance differences between subtasks are not merely an artifact of statistical noise. 

Circuit stability and the lack thereof also point to tangible ways that we can improve the model. For example, during training, circuit stability could possibly be improved through causal alignment~\cite{alignment,gupta_interpbench_2024}. Alternatively, stability could also be induced at inference via prompting. By explicitly breaking down a complex problem into simpler ones, we could encourage component/subtask reuse. We explore this latter possibility in Section~\ref{sec:case_study_prompting_effectiveness} through chain-of-thought prompting.

%% file: sections/boolean.tex
% !TEX root = ../acl_latex.tex

\section{Case Study: Boolean Expressions}\label{sec:case_study_boolean_expression_evaluation}
We now extend our analysis of circuit stability to a different task that also exhibits rich subtask structure: Boolean expression evaluation~\cite{suzgun-etal-2023-challenging}. Previously, we argued that circuit stability implies both length and compositional generalization. Here, we refine this perspective by showing that circuit non-equivalence or instability can also provide meaingful insights. Specifically, deviations in stability may indicate structural generalization~\cite{ye2021towards,he2024learning}. In other words, circuit stability is not simply a matter of ``more is better;'' rather, its desirability depends on how well it aligns with our prior knowledge of the task. 

\vspace{-0.1cm}
\subsection{Experimental Setup}
We use \texttt{phi-1.5}\footnote{A 1.5 billion parameter model with 128k circuit components.} due to its strong performance on logical and mathematical reasoning~\cite{li2023textbooksneediiphi15}. We follow the same evaluation setup as~\citet{srivastava2023beyond} for Boolean expression evaluation and prompt the model with $k=3$ few-shot examples. We construct partitions based on three independent variables: (1) expression length (number of words, e.g., \texttt{True and False} has length 3); (2) parenthetical depth (number of maximum nested parentheses, e.g., \texttt{(not (True))} has depth 2); and (3) the set of logical operators used (\texttt{not, and, or}). Expression length ranges from 1 to 9, and depth from 0 to 6. Circuit discovery details are largely the same as Section~\ref{sec:arithmetic-experimental-setup} and can be found in Appendix~\ref{app:circuit_discovery_details} and~\ref{app:reproducibility}.

\subsection{Circuit Instability and Generalization}

\noindent\textbf{Not Subtask.} Consider a Boolean expression that contains only the literals \texttt{True}, \texttt{False}, and the operator \texttt{not}. Since \texttt{not} is associative, adding and removing an arbitrary number of parentheses to any expression of this form should not change its ground-truth label. Thus, we expect a model that understands this axiom to apply the same circuit whether or not there are parentheticals in the expression. To test if this holds for \texttt{phi-1.5}, we first benchmark its circuit stability separately for sets of expressions with and without parentheses. 

\begin{figure}
\centering
\includegraphics[width=0.9\linewidth]{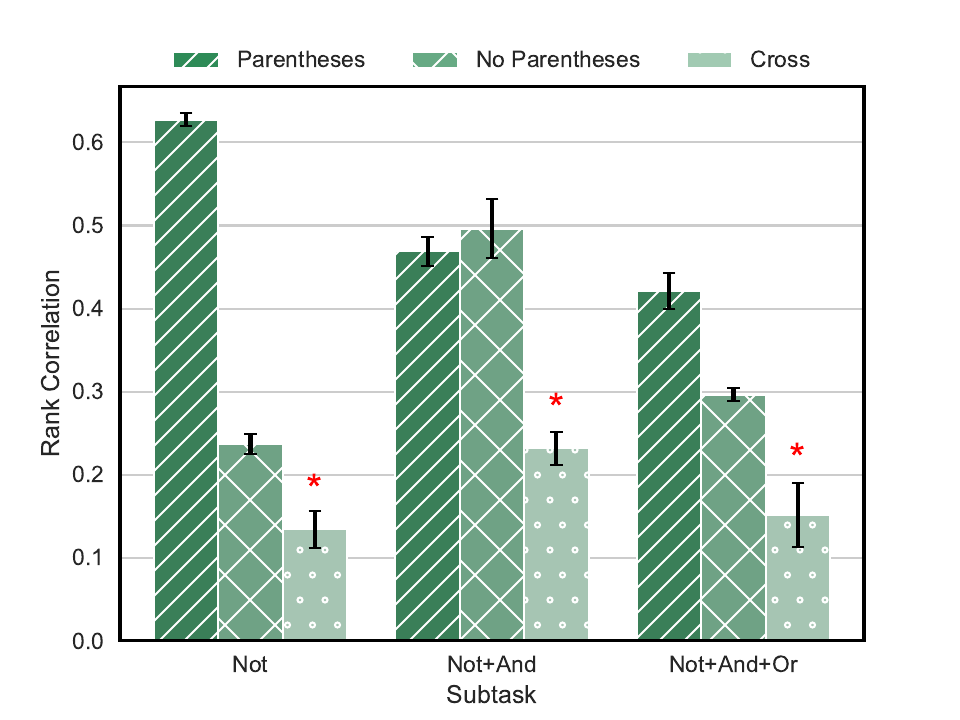}
\vspace{-0.15cm}
\caption{Circuit stability for \texttt{phi-1.5} within and across six subtasks. A permutation test is performed between parenthetical and non-parenthetical pairs of subtasks. ``{\color{red}{$\ast$}}'' denotes a statistically significant difference with a setting of $p< 0.05.$ The error bars denote a 95\% confidence interval.}
\label{fig:parenthetical-stability}
\vspace{-0.6cm}
\end{figure}

Concretely, for \texttt{not} expressions \textit{with} parentheses, we partition the input space by both parenthetical depth and expression length. In contrast, for \texttt{not} expressions \textit{without} parentheses, we partition only by expression length. The left two columns of the first bar group in Figure~\ref{fig:parenthetical-stability} illustrate \texttt{phi-1.5}'s respective circuit stability across these two partitions. 

The separately evaluated circuit stability of \texttt{phi-1.5} on each task  provides a baseline for computing $\alpha$-equivalence. If \texttt{phi-1.5} applies the same set of circuits across both parenthetical and non-parenthetical subtasks, then circuit stability should be consistent even as we permute the subtask soft circuits between these two groups. In particular, we expect this permutation not to cause circuit stability to drop below the minimum stability observed in across the two partitions. We find that \texttt{phi-1.5} applies statistically significant different circuits between these two tasks (see rightmost bar of first bar group in Figure~\ref{fig:parenthetical-stability}). 

\begin{figure}
\centering
\includegraphics[width=0.9\linewidth]{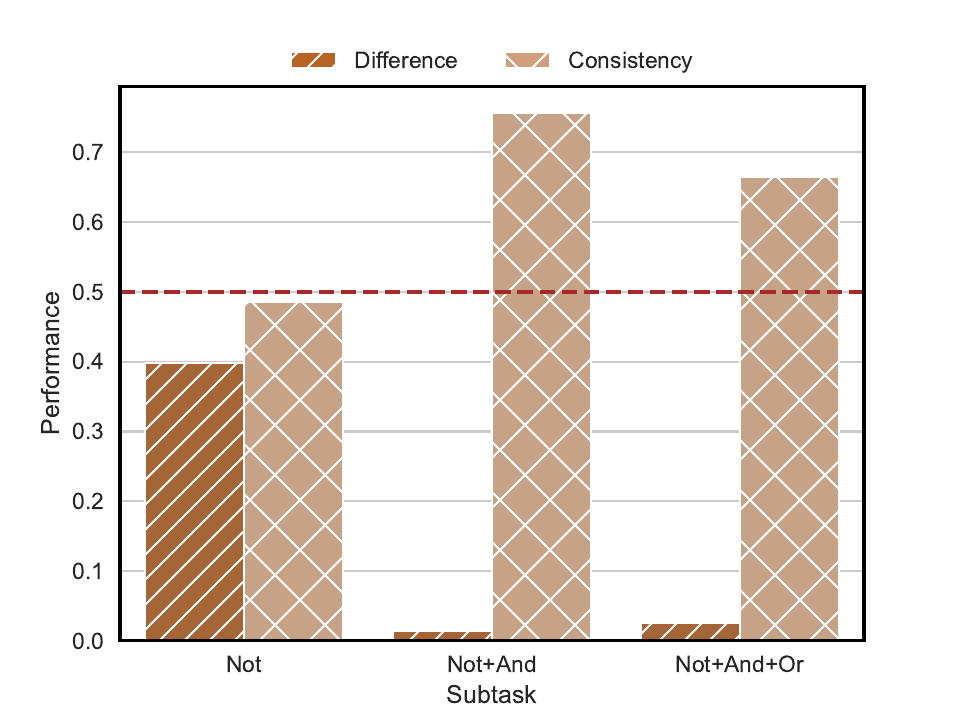}
\vspace{-0.15cm}
\caption{\textbf{(left columns)} The performance difference after adding parentheses. \textbf{(right columns)} The self-consistency of \texttt{phi-1.5} as a result of adding parentheticals. The dotted line denotes random chance of 0.5 for the consistency estimates.}
\label{fig:consist}
\vspace{-0.5cm}
\end{figure}

We hypothesize that this lack of circuit equivalence suggests that the model does not understand associativity in \texttt{not} evaluation. Indeed this is the case. Given any expression containing only \texttt{not}s and literals, after adding parentheses, \texttt{phi-1.5}'s performance decreases by 40\%. Further, the model is not self-consistent: adding parentheses to any expression causes the model to flip its prediction. This behavior is illustrated in the first bar group of Figure~\ref{fig:consist}.

\noindent\textbf{Not+And Subtask.} Now consider a Boolean expression with logical operators \texttt{not} and \texttt{and}. Adding and removing parentheses from this expression changes the order of evaluation. This may flip the ground-truth label\footnote{\texttt{(not False) and True != not False and True}}.  
Consequently, a model that respects operator precedence \textit{should} apply substantially different circuits across \texttt{not}+\texttt{and} expressions with and without parentheses. We apply the same experimental procedure as the previous \texttt{not} subtask. As shown in the second bar group of Figure~\ref{fig:parenthetical-stability}, we find that \texttt{phi-1.5} does employ different soft circuits across these partitions. Accordingly, in Figure~\ref{fig:consist}, we observe that \texttt{phi-1.5}'s performance stabilizes between subtasks with and without parentheses. Additionally, \texttt{phi-1.5} exhibits increased self-consistency. That is, for any particular expression, adding parentheses does not change the correctness of its prediction. 

\noindent\textbf{Not+And+Or Subtask.} Similar to the previous subtask, adding parentheses to an expression containing the operators \texttt{not}, \texttt{and}, and \texttt{or} alters the order of evaluation. As before, we observe that \texttt{phi-1.5}'s circuits align with our expectations (see third bar group of Figure~\ref{fig:parenthetical-stability}). As a result, we see consistent performance stability and increased self-consistency (see rightmost bar group in Figure~\ref{fig:consist}).

instability.

%% file: sections/prompting.tex
% !TEX root = ../acl_latex.tex
\vspace{-0.1cm}
\section{Case Study: Chain-of-Thought}\label{sec:case_study_prompting_effectiveness}
\vspace{-0.1cm}
In Section~\ref{sec:case_study_arithmetic_reasoning} and~\ref{sec:case_study_boolean_expression_evaluation}, we demonstrated that the stability of a model's circuit sheds light on its generalization. Now, we examine methods that tractably induce circuit stability. We hypothesize that chain-of-thought improves performance by promoting subtask decomposition and circuit component reuse~\cite{wei_chain--thought_2022}. As a result, we expect chain-of-thought to substantially improve circuit stability. Herein, we present some preliminary evidence for this claim. Unlike previous sections, we examine a task that is knowledge-based: sports understanding~\cite{suzgun-etal-2023-challenging}. Sports understanding is a binary classification task which presents models with sports statements and the model needs to decided whether they are true or false\footnote{For example, the model is presented with a statement like ``Santi Cazorla scored a touchdown.'' This statement is {\textit{false}} because Santi Cazorla is a soccer player and a ``touchdown'' is a part of American football and rugby~\cite{suzgun-etal-2023-challenging}.}. 

\begin{figure}
\centering
\includegraphics[width=0.9\linewidth]{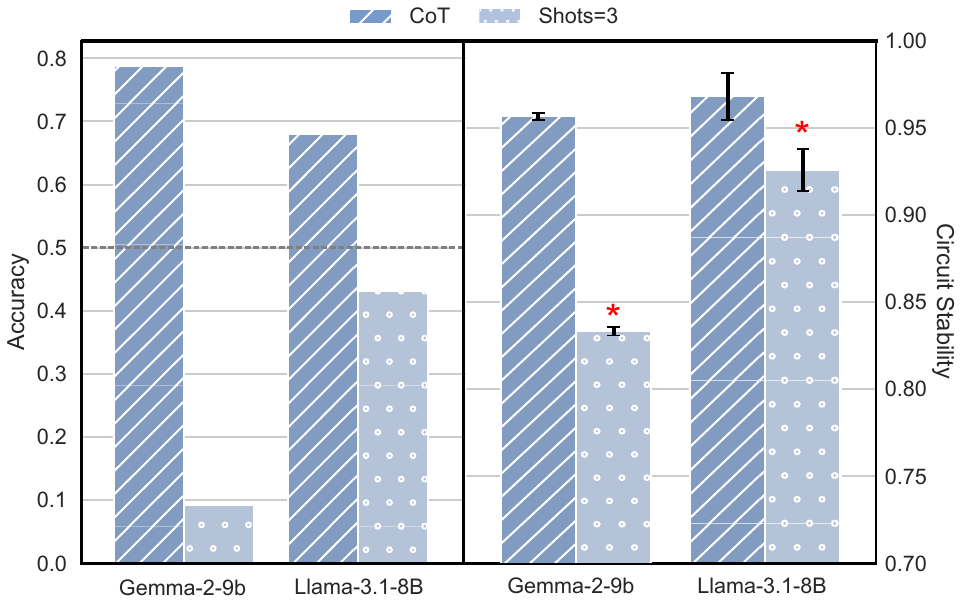}
\caption{\textbf{(left)} Exact string match accuracy of sports understanding task under chain-of-thought versus few-shot prompting. \textbf{(right)} Circuit stability across random partitions. We perform a two-sample $t$-test, ``{\color{red}{$\ast$}}'' denotes a significant difference ($p < 0.05)$.}
\label{fig:prompting-sports}
\vspace{-0.6cm}
\end{figure}

We employ both \texttt{Llama-3.1-8b} and \texttt{Gemma-2-9b}\footnote{8 billion parameter model containing over 1.5m circuit edges and 9 billion parameter model with over 720k circuit edges, respectively.} for this case study. Both models are sufficiently large to show significant performance improvements after prompting with chain-of-thought, see Figure~\ref{fig:prompting-sports}\textbf{(left)}. As before, model performance is also measured using exact string match accuracy. 

In contrast to our previous case studies, the subtask structure of this task much less apparent. As a result, we opt to construct subtasks simply by randomly partitioning the dataset into five disjoint cells. We compute the average circuit stability pairwise across these five cells before (we use few-shot prompting with $k=3$) and after chain-of-thought prompting. These results are shown in Figure~\ref{fig:prompting-sports}\textbf{(right)}. Across both models, we see that chain-of-thought significantly circuit stability. 

It should be noted that technically the partition strategy we employ herein is not creating true subtasks (see Definition~\ref{def:subtask}). This is because we are sampling from $\data$ a finite dataset \textit{first}, \textit{then} randomly partitioning the resulting dataset. As a result, the partitions we yield are i.i.d. with respect to $\data$. Thus, in some sense we are measuring the variance of the soft circuitry distribution before and after applying chain-of-thought. We could remedy this by fixing a partition strategy \textit{a priori} that with high probability induces similar substasks (in terms of transport distance). For example, we could partition subtasks based on the value of the fifth character of the input prompt. The connections between these two approaches could be explored more in future work.

%% file: sections/related.tex
% !TEX root = ../acl_latex.tex
\vspace{-0.1cm}
\section{Related Works}
\vspace{-0.1cm}
Most current work in mechanistic interpretability relies on \textit{ad hoc} interpretations tailored for a fixed task and model~\cite[\textit{inter alia}]{wang_interpretability_2022,stolfo-etal-2023-mechanistic,conmy_towards_2023,hanna_how_2023,arditi_refusal_2024,lee2024a,nikankin2025arithmetic}. As a result, generalizing these results into actionable insights remains difficult. To address this limitation, recent research has studied the dynamics of circuits and their mechanisms. This is also the focus of our paper. For example, works like \citet{nanda_progress_2022,zhong_clock_2023,humayun2024grokking,he2024learning,tigges_llm_2024} examine circuits across the training horizon which shed light on circuit formation and phenomena such as grokking~\cite{power_grokking_2022}. Further studies such as~\citet{lee2024a} seek to compare the tangible mechanistic differences before and after applying post-training methods like alignment. On the other hand, works such as~\citet{lieberum2023does,wu2023interpretability,merullo_circuit_2023} examine the change in a model's circuit with respect to scaling. 
Perhaps most similar to our line of work is the mechanistic interpretability of skill composition~\cite{arora2023theory,yu2024skillmix} which investigates how models combine learned skills to solve novel problems. Notably, \citet{chughtai_toy_2023,he2024learning} analyzes the emergence of skill composition in modular addition. However, they focus on small, toy models. It is unclear how their arguments and conclusions generalize to pretrained language models. In contrast, our framework of circuit stability provides a general characterization of circuits, their equivalence, and their stability---independent of the task or any specific model.

%% file: sections/discussion.tex
% !TEX root = ../acl_latex.tex
\vspace{-0.1cm}
\section{Conclusion and Discussion}\label{sec:discussion}
\vspace{-0.1cm}
In this paper, we introduce and formally define \textit{circuit stability} and \textit{equivalence} (Section~\ref{sec:circuit_stability}). We provide empirical evidence that circuit stability characterizes many key aspects of generalization and argue that this type of stability is actionable (Section~\ref{sec:case_study_arithmetic_reasoning} and~\ref{sec:case_study_boolean_expression_evaluation}). For example, it can be induced through prompting (Section~\ref{sec:case_study_prompting_effectiveness}). Our approach, based on methods from mechanistic interpretability, offers a step towards rigorously bridging the generality of models with their interpretability.

\textbf{Efficiency.} Our definition of a soft circuit (Definition~\ref{def:circuit}) involves continuous relaxation of the binary circuit function, $c: E_M \to \{0,1\}$. In most circuit discovery procedures, this relaxation, implemented via Equation~\ref{eq:activation-patching}, is necessary for tractability~\cite{conmy_towards_2023, nanda_attribution_2023, syed2023attribution}. As a result, many existing methods first perform this relaxation, then apply a greedy decoding strategy to extract the binary circuit function. Thus, our choice of both this continuous relaxation and Spearman's $\rho$ aligns with the standard practices of circuit discovery. Approximations for Definition~\ref{def:circuit} are also computationally efficient. Methods such as~\citet{syed2023attribution,hanna_have_2024} only require (a constant factor of) two forward and one backward pass of the model to estimate the entire soft circuitry.

\textbf{Occam's Razor.} Crucially, soft circuitry does not account for higher level algorithmic similarities. That is, we could have different soft circuits which also correspond to distinct hard circuits, but algorithmically they implement the same procedure~\cite{olsson2022context,merullo_circuit_2023}. We do not see this as a limitation of the work but rather a feature of our mathematical framework. From a learning-theoretic perspective, duplicate mechanisms necessarily imply longer minimum description lengths. In turn, this leads to looser generalization bounds~\cite[\textit{inter alia}]{hansen2001model,sefidgaran_minimum_2023}. More informally, if we subscribe to the principle of Occam's Razor~\cite{blumer_occams_1987,shalev-shwartz_understanding_2014} then we would also prefer a model with less duplicate mechanisms. As a result, our analyses implicitly take into account these learning-theoretic constructs. Lastly, if one was truly concerned with algorithmic differences, then $K$ could be augmented using metrics from causal abstraction~\cite[\textit{inter alia}]{beckers_abstracting_2019,abstraction,otsuka_equivalence_2022,alignment}. However, this introduces additional complexities that we leave for future work. 

\textbf{Finding Partitions.} Throughout Sections~\ref{sec:case_study_arithmetic_reasoning},~\ref{sec:case_study_boolean_expression_evaluation}, we leverage prior knowledge about the task to construct partitions of interest. But, for more complex tasks requiring an intricate composition of skills, the appropriate partitions may not be obvious. In Section~\ref{sec:case_study_prompting_effectiveness}, we demonstrate that this does not hinder the practicality of circuit stability. Even randomly chosen partitions can yield meaningful insights. Alternatively, since soft circuitry is an expectation, we could have also sought to characterize circuit stability through the asymptotic variance of its limiting distribution. Further, we hypothesize that if any of the partitions contain cells that are $\eps$-representative~\cite{shalev-shwartz_understanding_2014} with respect to both $L$ and $\data$ in Equation~\ref{eq:activation-patching}, then they should lead to a sharp characterization of the model's performance. Investigating this connection could form the basis of interesting future work and reduce reliance on constructing partitions manually.

%% file: appendix/limitations.tex
% !TEX root = ../acl_latex.tex

\section{Limitations}\label{app:limitations}

Our empirical case studies in Sections~\ref{sec:case_study_arithmetic_reasoning},~\ref{sec:case_study_boolean_expression_evaluation}, and~\ref{sec:case_study_prompting_effectiveness} are fairly limited in terms of the tasks and models we benchmark. We hope that these preliminary results will give the larger research community a taste of how circuit stability can be used and leave these extensions for future work. 

Another limitation of the work is the choice of circuit abstraction~\cite{vilas_position_2024}. In Section~\ref{sec:background}, we loosely defined the model's circuit as a subgraph of its computational graph. However, there are many ways this computational graph could be specified. On one extreme, there exists the trivial computation graph: an input and output node with a single edge representing the entire function. On the other extreme, we could define the graph as a trace of the compiled machine code. This is also a challenge that mechanistic interpretability faces. It is unclear how circuit stability would react to these different levels of abstraction. Perhaps future theoretical analyses of circuit stability could take this into account by involving $|V^M|$ into its bounds.

%% file: appendix/beyond.tex
% !TEX root = ../acl_latex.tex

\section{Circuit Stability Beyond Transformer Language Models}\label{app:beyond}
In this paper, we focused our case studies on Transformer language models, as their circuits are well-studied in the the literature~\cite{elhage_mathematical_2021,olsson2022context,wang_interpretability_2022, hanna_how_2023}. However, the concept of the circuit stability can be extended to any neural network as long as it admits a consistent computational graph. In this section, we present three distinct neural architectures associated with applications outside of the language domain and demonstrate that even in these cases, circuit stability is a well-defined notion. In this way, circuit stability is a modality-independent concept. An interesting line of future work could be to extend this framework to study the circuits and generalization of vision-language models~\cite{bordes2024introduction}.

\noindent\textbf{Fully-Connected Neural Networks.} A fully-connected network consists of successive linear layers with activation functions: $M = f_n \circ \ldots \circ f_1$ where $f_i : \R^{d_i} \to \R^{d_{i+1}}$. Here, $d_1, d_{n+1}$ are the input and output dimensions. For given activation functions $\sigma_i : \R \to \R$, a single fully-connected layer is defined as $f_i(x) = \sigma_i(W_ix)$, where $W_i \in \R^{d_{i+1} \times d_i}$ and $\sigma_i$ is applied element-wise. There are many ways to construct a valid computational graph upon this architecture. For brevity, we only show the most intuitive construction: a node in the $i$\textsuperscript{th} layer is simply an entry in $x$. Specifically, we can decompose the $j$\textsuperscript{th} entry in layer $i$ as ${(\sigma_i(W_i x))}_j = \sigma_i\left(\sum_{k=1}^{d_i} W_{jk} x_k \right)$. Note also that each entry in the input should also be assigned a node. Therefore, this node forms edges with all nodes in the previous layer. 

\noindent\textbf{Convolutional Neural Networks.} We take a similar approach to the decomposition of fully-connected neural networks. For simplicity, we only consider the case of a single 2d-convolution layer, a convolutional network with higher dimensions or more layers can be derived inductively. We directly lift this description of a convolutional layer from~\cite{sun2024achieving}. Let $f : \R^{c\times h \times w} \rightarrow \R^{c'\times h' \times w'}$. Suppose that this layer is parameterized by kernels $W_i \in \R^{k\times k}$ for $1 \leq i \leq  c'$ and some $k \in \N$ as well as a bias $b \in \R^{c'}$. Then, it follows that
\begin{align*}
f(x)_j = \left(\bm{1}_{h'\times w'}b_j + \sum_{i=1}^c W_j \ast x[i,:,:]\right),
\end{align*}
for $1 \leq j \leq c'$ where $f(x)_j \in \R^{h'\times w'}$ for $h',w'$ being the resulting dimension after convolution with a $k\times k$ kernel. Here, $\bm{1}_{h'\times w'} \in \R^{h'\times w'}$ is a one matrix. Then, one convolutional layer needs $c'h'w' + chw$ nodes: one for each input and output coordinate. A directed edge is drawn from all input nodes to each output node\footnote{This may not be the tightest graph that we can build in terms of the number of edges we need to construct. However, after performing any ablations, we should expect those edges that are non-tight to contribute a score of 0.}.

\noindent\textbf{State-Space Language Models.} A state-space model processes input tokens sequentially. Each token, $x_t$ is processed in constant time~\cite{ssms} via a hidden state $s_t$ such that 
\begin{align*}
s_{t+1} &= \bm{A}s_t + \bm{B}x_t, \\
y_t &= \bm{C}s_t + \bm{D}x_t.
\end{align*}
Thus, we can the computational at each time-step to essentially be a combination of two fully-connected neural networks. We can specify a well-defined computational graph by unrolling the recursive equations above. We define each $x_t$ for all $t$ as a node and each entry of $s_t$ to also be a node. The, the edges are given by the same schema we used to define the fully-connected layers.

\noindent\textbf{Putting it All Together.} These examples demonstrate that as long as we can specify a computational graph, the circuit of a model is well-defined. Therefore, the existence of circuit stability as well as its implications depends largely on the task distribution (Definition~\ref{def:task}), its subtask distributions (Definition~\ref{def:subtask}), and the granularity of the computational graph. To this last point, \textit{a priori} it is unclear what the correct level of abstraction one should impose on the nodes and edges. On one end, we could define the nodes as the input/output of each FLOP of compute. However, though this might provide lots of insight, the computation of its circuit (Definition~\ref{def:circuit}) would be intractable. On the other end, we could define the computational graph as a single node with no edges. Though this is computationally more favorable, it yields no insights. The subfield of causal abstraction addresses some of these issues and we refer the reader to~\citet{abstraction}.

%% file: appendix/patching.tex
% !TEX root = ../acl_latex.tex

\section{Circuit Discovery Details}\label{app:circuit_discovery_details}
For all of our circuit discovery experiments, we perform \textit{attribution
patching}~\cite{nanda_attribution_2023}. Attribution patching is a linear
approximation to \textit{activation patching} also called causal 
mediation analysis~\cite{heimersheim_how_2024}. First introduced by
\citet{vig_investigating_2020} and extended upon by 
\citet{meng_locating_2022, wang_interpretability_2022,
conmy_towards_2023}, activation patching
seeks to determine the effect of a single neural component on the entire model's
output, for some fixed task. Then, by isolating all such components, 
we have effectively found the subcircuit responsible for the model's behavior on
this specific task. More formally, let $L(\cdot)$ be a 
function that maps a model's output into a scalar, generically this could be
some loss function. 
This is the \textit{patching metric}. For notational brevity, we omit $L$'s dependency on the ground truth label. Let $c$ be an
arbitrary neural network component that we wish to patch, and also denote
by $x$ an arbitrary input to our model $M$. Using \citeposs{
pearl2009causality} notion of do-calculus, activation patching can be written
as 
\begin{equation}\label{eq:activation-patching}
\textrm{Patch}_c(x, c^\star) \coloneqq
L(M^{\doc(c = c^\star)}(x)) - L(M(x)),
\end{equation}
where $c^\star$ is a \textit{counterfactual} output of activation $c$ that
we patch in. In natural language, $\text{Patch}_c(x, c^\star)$ can be expressed
as ``if we replace only the output of component $c$ with $c^\star$, how will
the model now behave?'' 
We encourage the reader to refer to 
\citet{heimersheim_how_2024} for a detailed introduction to activation
patching and instead briefly explain attribution patching, our design choices,
as well as our extensions to the multi-token setting.

One drawback of activation patching is its computational cost. Consider a dataset
of $n$ data points and a model with $k$ neural components we are interested in
patching, activation patching would require $\bigO(nk)$ forward passes. This becomes
prohibitively expensive when $k \gg 1$~\cite{kramar2024atp}. Thus, 
using Equation~\ref{eq:activation-patching} as a jumping off point, attribution
patching seeks to make activation patching more efficient. Consider a first-order
Taylor series approximation of $L$ around $c$ assuming that 
$c^\star \approx c'$, where $c'$ is the unpatched activation of $c$ on $x$. Then,
it follows that 
\begin{equation}\label{eq:attribution-patching}
\textrm{Patch}_c(x, c^\star) \approx
(c' - c^\star) \nabla_c L(x).
\end{equation}
Crucially, \citet{nanda_attribution_2023,syed2023attribution} argue that this new
metric can be computed in two forward passes and one backward pass for all components. Thus, we only
require $\bigO(n)$ forward and backward passes. Throughout the paper, we use a 
variant of attribution patching
introduced by~\cite{hanna_have_2024} called \textit{edge attribution patching} with
\textit{integrated gradients} (EAP-IG). In short, EAP-IG deduces a more accurate
approximation to activation patching than vanilla attribution patching in
Equation~\ref{eq:attribution-patching}. EA-IG operates by computing a path integral
from $c' \to c^\star$:
\begin{gather}
\Delta_{c', c^\ast} \int_0^1 
\frac{\partial L}{\partial c'}M^{\doc(c =  
\Delta_{c', -\alpha(c^\star - c')})}(x)~d\alpha\label{eq:eap-ig} \\ \approx
\Delta_{c', c^\ast} \frac{1}{m}\sum_{k=1}^m 
\frac{\partial L}{\partial c'} M^{\doc(c = \Delta_{c', -k(c^\star - c')/m })}(x),
\label{eq:empirical-integral-eap}
\end{gather}
where $\Delta_{c', c} = c' - c$ and $m$ is a hyperparameter representing the number of steps
to approximate the integral. Equation~\ref{eq:empirical-integral-eap} can be understood
as a Monte-Carlo estimate of the integral in Equation~\ref{eq:eap-ig}. 
Generally, for such estimates to be accurate it requires $m \gg 1$, potentially
on the order of $\sim 10^5$. However, empirically \citet{hanna_have_2024} finds
even $m=5$ works quite well. These hyperparameters are adopted for all 
of the circuit experiments.

\subsection{Patching Multi-token Tasks}

\begin{table*}[h]
\centering
\begin{tabular}{lll}
\hline Feature & Next-token Patching & Joint-token Patching
\\ \hline 
Granularity & Token & Sequence \\ 
Focus & Localized, stepwise effects & Global multi-token coherence \\ 
Computational Cost & Can compute exactly & High; requires approximations \\
Insight & Fine-grained, task-specific behavior & Broader token dependencies \\ \hline
\end{tabular}
\caption{Features of next-token patching versus joint-token patching. The 
former can be seen as grokking the local token-level behavior of the model
compared to the latter which can be viewed as uncovering global token
dependencies.}
\label{table:patching-schemas}
\end{table*}

Herein, we detail circuit discovery in the case of
multitoken tasks. To the best of our knowledge, almost all
of the mechanistic interpretability literature deals with tasks
that require only a single token output. So, our methods represent one attempt to generalize these existing approaches. We present
two distinct approaches, and briefly discuss their interpretations (Table~\ref{table:patching-schemas}).

Denote by $p$ some prompt. Let $t_1,t_2,\ldots,t_n$ be new tokens generated
autoregressively by some model $M$. We fix $n \geq 1$ and focus most of
our analysis on the case where $n=2$ as any finite $n$ can be derived
inductively. Also denote by $\P$ the probability distribution over
$t_1,t_2,\ldots,t_n$ conditioned on some prompt $p$ induced by the model $M$
~\cite{cotterell2023formal,du-etal-2023-measure}. Let $\P^{\doc}$ be the model
after intervention by one of the methods described previously. 

\noindent\textbf{Next-token patching.} Let 
$t_1^\star, t_2^\star, \ldots, t^\star_n$ be the expected output for a given
prompt $p$. For a fixed prompt, next-token patching defines the
following patching metric
\begin{gather}
\textrm{NextToken}(p) \coloneqq \kl( \P[t_1 | p] \parallel 
\P^{\doc}[t_1 | p] ) + \\\sum_{i=2}^n
\kl\left(\P[t_i | p, t_{[:i]}^\star] \parallel 
\P^{\doc}[t_i | p, t_{[:i]}^\star]\right).\label{eq:next-token-patching}
\end{gather}
$\kl(\P \parallel \P^{\doc})$ denotes the KL-divergence between
$\P$ and $\P^{\doc}$.
Essentially, Equation~\ref{eq:next-token-patching} measures the effect of 
patching any given component on model $M$'s next word prediction 
ability. Specifically, this metric captures a ``local property'' of
$M$ since in each summand, we assume that $M$ has previously generated
the correct tokens. To compute the patching metric across the entire
task distribution, we simply take the expectation over $p$: 
$\E_{p}[\textrm{NextToken}(p)]$.

\noindent\textbf{Joint-token patching.} As discussed previously,
a language model can be thought of as a measure over the space of all
sentences: $\P$. Joint-token patching directly measures the difference 
between the measure $\P$ (induced by $M$) and $\P^{\doc}$ (induced by
an intervention on $M$) over the joint distribution of all $n$-tokens.
Concretely, for a fixed prompt $p$, 
\begin{align*}
\textrm{JointToken}(p) \coloneqq \kl(\P[t_1\ldots t_n | p] 
\parallel \P^{\doc}[t_1\ldots t_n | p]).
\end{align*}
By expanding the definition of $\kl(\cdot \parallel \cdot)$, it is
easy to see that for $n=2$:
\begin{align*}
\textrm{JointToken}(p) = \kl(\P[t_1 | p] \parallel \P^{\doc}[t_1 | p])
+ \\\E_{t_1 \sim \P[t_1 | p]}\left[\kl(\P[t_2 | p, t_1] 
\parallel \P^{\doc}[t_2 |
p, t_1])\right].
\end{align*}
Note the expectation in the second summand. This is with respect
to the token $t_1$ generated by the model without any intervention.
The computation of this expectation is hard especially if $n$ is large. 

In both next-token and joint-token patching KL-divergence is used. 
This follows from the recommendation and positive results of~\citet{conmy_towards_2023}, 
but one can technically use any other preferred metric. The key idea being that the metric should capture the performance decrease of the model after ablating an individual edge.

\subsection{Noisy-to-Clean Patching}
In Section~\ref{sec:case_study_arithmetic_reasoning} and~\ref{sec:case_study_boolean_expression_evaluation}, we perform noisy-to-clean patching. That is, for a model $M$, we run the model on a given example $x$ and cache all of the activations across all of the edges. Then, we take another example $x'$ which is associated with a different ground truth example. While the model runs on $x'$, we patch in activations from $x$ and check to see how much performance decreases with respect to the label associated with $x'$. In the case of attribution patching, we apply the same Taylor-expansion as above. Noisy-to-clean activation gives us a picture of the necessary components of the model. 
\subsection{Clean-to-Clean Patching}
\vspace{-0.1cm}
In cases where there is not necessarily a ground-truth label, noisy-to-clean patching is not well-defined. Herein, we propose a new type of patching which we call clean-to-clean patching. This is applied in Section~\ref{sec:case_study_prompting_effectiveness} and generally useful for circuit discovery where the output is some open-form generation for autoregressive language models. Let $x$ be the input and $x^\star$ be the model's sampled response of $x$. We first run the model on $x$ with padding tokens such that $x$ and $x^\star$ have the same length. Similar to the procedure above, we cache all of the intermediate activations. Then, we run the model on $x^\star$ and patch in the appropriate activations from this ``blanked'' out $x$. The idea here is that we are finding the necessary components that exactly recover the model's response on this input, where the baseline is if the model had not generated anything at all. In the case of attribution patching, we apply the same Taylor-expansion as above. 

%% file: appendix/napkin.tex
% !TEX root = ../acl_latex.tex

\section{Within-Task Regressions}\label{app:napkin_calcuations}

\begin{figure}
\centering
\includegraphics[width=\linewidth]{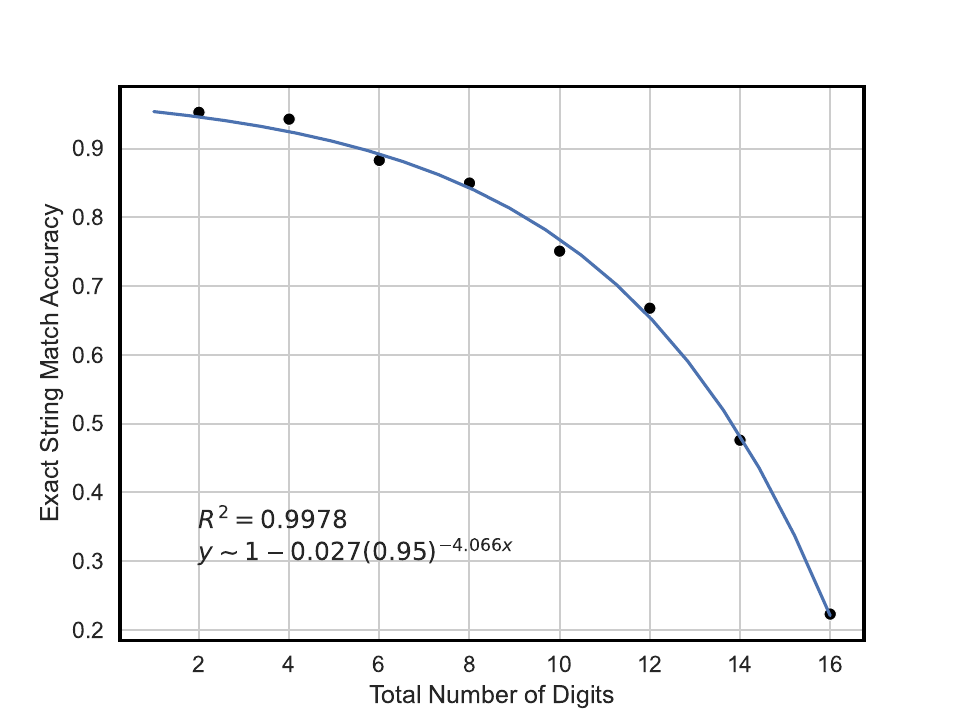}
\caption{Exponential regression on exact string accuracy for equal-digit subtasks. That is, $o_1=o_2$. The $x$-axis of the regression is the sum of the number of digits in both operands.}
\label{fig:diag-regress}
\vspace{-0.5cm}
\end{figure}

\begin{figure}
\centering
\includegraphics[width=\linewidth]{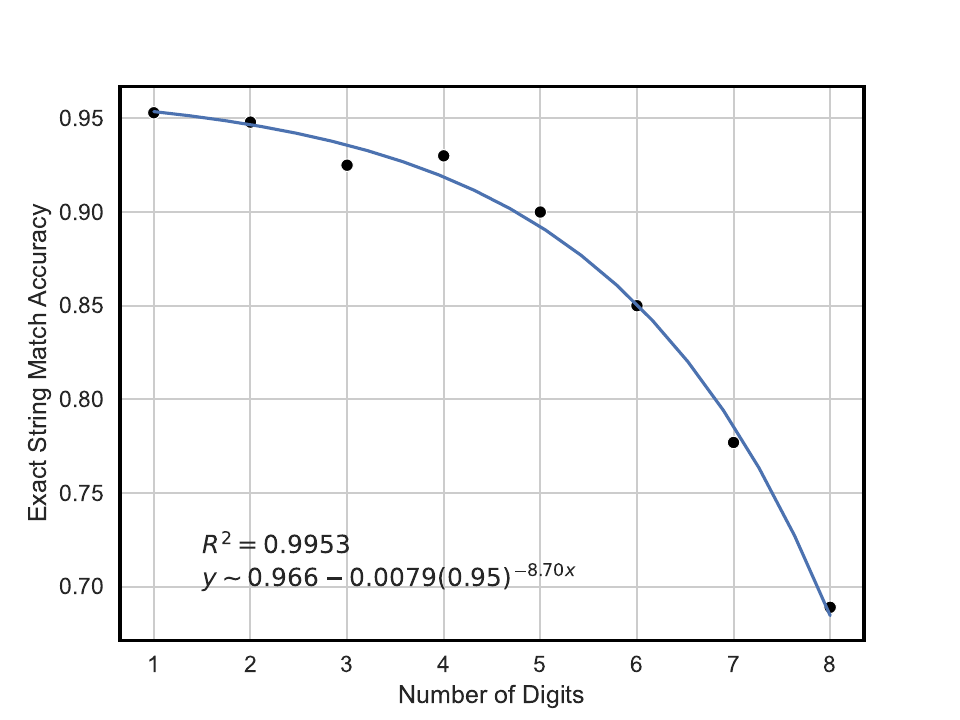}
\caption{Exponential regression on exact string accuracy for one-digit subtasks. That is $o_2 = 1$. The $x$-axis of the regression is simply the number of digits in $o_1$ as $o_2$ is constant. }
\label{fig:side-regress}
\vspace{-0.5cm}
\end{figure}

In Section~\ref{sec:case_study_arithmetic_reasoning}, we argued that within an $\alpha$-equivalent circuit family, performance decreases predictability. Herein, we perform a regression analysis to analyze this hypothesis. Specifically, we examine whether the subtask performances shown in Figure~\ref{fig:benchmark} can be predicted through an exponential regression using the performance of the $(1,1)$ subtask. We perform an exponential regression since we expect error to be compounded. That is, for a $(2,2)$ task, at least 4 $(1,1)$ subtasks need to be computed. Thus, the error should be compounding on the order of $0.95^4$. Likewise, a subtask that requires $n$, $(1,1)$ subtask decompositions should require incur error on the order of $\bigO(0.95^n)$. Therefore, consider a regression of the form $y = b - c(0.95)^{ax}$, where $a,b,c$ are learnable parameters and $x$ is the total number of digits across both operands, and $ax$ is the total number of subtask decompositions. 
These regression results are shown in Figure~\ref{fig:diag-regress} and Figure~\ref{fig:side-regress}. We find that a strong $R^2$ is observed $>0.99$.

%% file: appendix/reproduction.tex
% !TEX root = ../acl_latex.tex

\section{Reproducibility}\label{app:reproducibility}

Throughout the paper, we do not perform any finetuning or training. Rather, we directly evaluate the pretrained models.
All of our experiments were conducted on two NVIDIA A100 80GB GPUs. Our codebase including the implementations of the proposed algorithms and figures can be found at \url{https://github.com/alansun17904/circuit-stability}. The models and their weights used as case study throughout the paper are loaded directly from the \texttt{transformer\_lens} package, its documentation can be found at \url{https://transformerlensorg.github.io/TransformerLens/index.html}. We use all of the default hyperparameters and settings of the package.

%% file: acl_latex.bbl
\begin{thebibliography}{64}
\providecommand{\natexlab}[1]{#1}

\bibitem[{Adolfi et~al.(2025)Adolfi, Vilas, and Wareham}]{adolfi2025the}
Federico Adolfi, Martina~G. Vilas, and Todd Wareham. 2025.
\newblock \href {https://openreview.net/forum?id=QogcGNXJVw} {The computational complexity of circuit discovery for inner interpretability}.
\newblock In \emph{The Thirteenth International Conference on Learning Representations}.

\bibitem[{Arditi et~al.(2024)Arditi, Obeso, Syed, Paleka, Panickssery, Gurnee, and Nanda}]{arditi_refusal_2024}
Andy Arditi, Oscar Obeso, Aaquib Syed, Daniel Paleka, Nina Panickssery, Wes Gurnee, and Neel Nanda. 2024.
\newblock \href {https://proceedings.neurips.cc/paper_files/paper/2024/file/f545448535dfde4f9786555403ab7c49-Paper-Conference.pdf} {Refusal in {Language} {Models} {Is} {Mediated} by a {Single} {Direction}}.
\newblock In \emph{Advances in {Neural} {Information} {Processing} {Systems}}, volume~37, pages 136037--136083. Curran Associates, Inc.

\bibitem[{Arora and Goyal(2023)}]{arora2023theory}
Sanjeev Arora and Anirudh Goyal. 2023.
\newblock A theory for emergence of complex skills in language models.
\newblock \emph{arXiv preprint arXiv:2307.15936}.

\bibitem[{Beckers and Halpern(2019)}]{beckers_abstracting_2019}
Sander Beckers and Joseph~Y Halpern. 2019.
\newblock Abstracting causal models.
\newblock In \emph{Proceedings of the {AAAI} {Conference} on {Artificial} {Intelligence}}, volume~33, pages 2678--2685.
\newblock Issue: 01.

\bibitem[{Bhaskar et~al.(2024)Bhaskar, Wettig, Friedman, and Chen}]{bhaskar_finding_2024}
Adithya Bhaskar, Alexander Wettig, Dan Friedman, and Danqi Chen. 2024.
\newblock \href {https://proceedings.neurips.cc/paper_files/paper/2024/file/20fdaf67581e6d7157376d1ed584040a-Paper-Conference.pdf} {Finding {Transformer} {Circuits} {With} {Edge} {Pruning}}.
\newblock In \emph{Advances in {Neural} {Information} {Processing} {Systems}}, volume~37, pages 18506--18534. Curran Associates, Inc.

\bibitem[{Blumer et~al.(1987)Blumer, Ehrenfeucht, Haussler, and Warmuth}]{blumer_occams_1987}
Anselm Blumer, Andrzej Ehrenfeucht, David Haussler, and Manfred~K. Warmuth. 1987.
\newblock \href {https://doi.org/10.1016/0020-0190(87)90114-1} {Occam's {Razor}}.
\newblock \emph{Information Processing Letters}, 24(6):377--380.

\bibitem[{Bordes et~al.(2024)Bordes, Pang, Ajay, Li, Bardes, Petryk, Ma{\~n}as, Lin, Mahmoud, Jayaraman et~al.}]{bordes2024introduction}
Florian Bordes, Richard~Yuanzhe Pang, Anurag Ajay, Alexander~C Li, Adrien Bardes, Suzanne Petryk, Oscar Ma{\~n}as, Zhiqiu Lin, Anas Mahmoud, Bargav Jayaraman, et~al. 2024.
\newblock An introduction to vision-language modeling.
\newblock \emph{arXiv preprint arXiv:2405.17247}.

\bibitem[{Cho et~al.(2025)Cho, Cha, Bhojanapalli, and Yun}]{cho2025arithmetic}
Hanseul Cho, Jaeyoung Cha, Srinadh Bhojanapalli, and Chulhee Yun. 2025.
\newblock \href {https://openreview.net/forum?id=eIgGesYKLG} {Arithmetic transformers can length-generalize in both operand length and count}.
\newblock In \emph{The Thirteenth International Conference on Learning Representations}.

\bibitem[{Chughtai et~al.(2023)Chughtai, Chan, and Nanda}]{chughtai_toy_2023}
Bilal Chughtai, Lawrence Chan, and Neel Nanda. 2023.
\newblock \href {https://arxiv.org/pdf/2302.03025} {A {Toy} {Model} of {Universality}: {Reverse} {Engineering} how {Networks} {Learn} {Group} {Operations}}.
\newblock In \emph{Proceedings of the 40th {International} {Conference} on {Machine} {Learning}}.

\bibitem[{Conmy et~al.(2023)Conmy, Parker-Mavor~N., Lynch, Heimersheim, and Alonso-Garriga}]{conmy_towards_2023}
Arthur Conmy, Augustine Parker-Mavor~N., Aengus Lynch, Stefan Heimersheim, and Adria Alonso-Garriga. 2023.
\newblock \href {https://arxiv.org/abs/2304.14997} {Towards {Automated} {Circuit} {Discovery} for {Mechanistic} {Interpretability}}.
\newblock In \emph{Thirty-{Seventh} {Conference} on {Neural} {Information} {Processing} {Systems}}.

\bibitem[{Cotterell et~al.(2023)Cotterell, Svete, Meister, Liu, and Du}]{cotterell2023formal}
Ryan Cotterell, Anej Svete, Clara Meister, Tianyu Liu, and Li~Du. 2023.
\newblock Formal aspects of language modeling.
\newblock \emph{arXiv preprint arXiv:2311.04329}.

\bibitem[{Cousot and Cousot(1977)}]{cousot1977abstract}
Patrick Cousot and Radhia Cousot. 1977.
\newblock \href {https://doi.org/10.1145/512950.512973} {Abstract interpretation: a unified lattice model for static analysis of programs by construction or approximation of fixpoints}.
\newblock In \emph{Proceedings of the 4th ACM SIGACT-SIGPLAN Symposium on Principles of Programming Languages}, POPL '77, page 238–252, New York, NY, USA. Association for Computing Machinery.

\bibitem[{Du et~al.(2023)Du, Torroba~Hennigen, Pimentel, Meister, Eisner, and Cotterell}]{du-etal-2023-measure}
Li~Du, Lucas Torroba~Hennigen, Tiago Pimentel, Clara Meister, Jason Eisner, and Ryan Cotterell. 2023.
\newblock \href {https://doi.org/10.18653/v1/2023.acl-long.543} {A measure-theoretic characterization of tight language models}.
\newblock In \emph{Proceedings of the 61st Annual Meeting of the Association for Computational Linguistics (Volume 1: Long Papers)}, pages 9744--9770, Toronto, Canada. Association for Computational Linguistics.

\bibitem[{Elhage et~al.(2021)Elhage, Nanda, Olsson, Henighan, Joseph, Mann, Askell, Bai, Chen, Conerly, {DasSarma, Nova}, Drain, Ganguli, Hatfield-Dodds, Hernandez, Jones, Kernion, Lovitt, Ndousse, Amodei, Brown, Clark, Kaplan, McCandlish, and Olah}]{elhage_mathematical_2021}
Nelson Elhage, Neel Nanda, Catherine Olsson, Tom Henighan, Nicholas Joseph, Ben Mann, Amanda Askell, Yuntao Bai, Anna Chen, Tom Conerly, {DasSarma, Nova}, Dawn Drain, Deep Ganguli, Zac Hatfield-Dodds, Danny Hernandez, Andy Jones, Jackson Kernion, Liane Lovitt, Kamal Ndousse, Dario Amodei, Tom Brown, Jack Clark, Jared Kaplan, Sam McCandlish, and Chris Olah. 2021.
\newblock \href {https://transformer-circuits.pub/2021/framework/index.html} {A {Mathematical} {Framework} for {Transformer} {Circuits}}.

\bibitem[{Geiger et~al.(2021)Geiger, Lu, Icard, and Potts}]{abstraction}
Atticus Geiger, Hanson Lu, Thomas Icard, and Christopher Potts. 2021.
\newblock \href {https://proceedings.neurips.cc/paper_files/paper/2021/file/4f5c422f4d49a5a807eda27434231040-Paper.pdf} {Causal abstractions of neural networks}.
\newblock In \emph{Advances in Neural Information Processing Systems}, volume~34, pages 9574--9586. Curran Associates, Inc.

\bibitem[{Geiger et~al.(2024)Geiger, Wu, Potts, Icard, and Goodman}]{alignment}
Atticus Geiger, Zhengxuan Wu, Christopher Potts, Thomas Icard, and Noah Goodman. 2024.
\newblock \href {https://proceedings.mlr.press/v236/geiger24a.html} {Finding alignments between interpretable causal variables and distributed neural representations}.
\newblock In \emph{Proceedings of the Third Conference on Causal Learning and Reasoning}, volume 236 of \emph{Proceedings of Machine Learning Research}, pages 160--187. PMLR.

\bibitem[{Glazer et~al.(2024)Glazer, Erdil, Besiroglu, Chicharro, Chen, Gunning, Olsson, Denain, Ho, Santos et~al.}]{glazer2024frontiermath}
Elliot Glazer, Ege Erdil, Tamay Besiroglu, Diego Chicharro, Evan Chen, Alex Gunning, Caroline~Falkman Olsson, Jean-Stanislas Denain, Anson Ho, Emily de~Oliveira Santos, et~al. 2024.
\newblock Frontiermath: A benchmark for evaluating advanced mathematical reasoning in ai.
\newblock \emph{arXiv preprint arXiv:2411.04872}.

\bibitem[{Gu et~al.(2021)Gu, Johnson, Goel, Saab, Dao, Rudra, and R\'{e}}]{ssms}
Albert Gu, Isys Johnson, Karan Goel, Khaled Saab, Tri Dao, Atri Rudra, and Christopher R\'{e}. 2021.
\newblock \href {https://proceedings.neurips.cc/paper_files/paper/2021/file/05546b0e38ab9175cd905eebcc6ebb76-Paper.pdf} {Combining recurrent, convolutional, and continuous-time models with linear state space layers}.
\newblock In \emph{Advances in Neural Information Processing Systems}, volume~34, pages 572--585. Curran Associates, Inc.

\bibitem[{Gupta et~al.(2024)Gupta, Arcuschin, Kwa, and Garriga-Alonso}]{gupta_interpbench_2024}
Rohan Gupta, Iván Arcuschin, Thomas Kwa, and Adrià Garriga-Alonso. 2024.
\newblock \href {https://proceedings.neurips.cc/paper_files/paper/2024/file/a8f7d43ae092d9a5295775eb17f3f4f7-Paper-Datasets_and_Benchmarks_Track.pdf} {{InterpBench}: {Semi}-{Synthetic} {Transformers} for {Evaluating} {Mechanistic} {Interpretability} {Techniques}}.
\newblock In \emph{Advances in {Neural} {Information} {Processing} {Systems}}, volume~37, pages 92922--92951. Curran Associates, Inc.

\bibitem[{Hanna et~al.(2023)Hanna, Liu, and Variengien}]{hanna_how_2023}
Michael Hanna, Ollie Liu, and Alexandre Variengien. 2023.
\newblock \href {https://openreview.net/forum?id=p4PckNQR8k} {How does {GPT}-2 compute greater-than?: {Interpreting} mathematical abilities in a pre-trained language model}.
\newblock In \emph{Thirty-seventh {Conference} on {Neural} {Information} {Processing} {Systems}}.

\bibitem[{Hanna et~al.(2024)Hanna, Pezzelle, and Belinkov}]{hanna_have_2024}
Michael Hanna, Sandro Pezzelle, and Yonatan Belinkov. 2024.
\newblock \href {https://openreview.net/forum?id=TZ0CCGDcuT} {Have {Faith} in {Faithfulness}: {Going} {Beyond} {Circuit} {Overlap} {When} {Finding} {Model} {Mechanisms}}.
\newblock In \emph{First {Conference} on {Language} {Modeling}}.

\bibitem[{Hansen and and(2001)}]{hansen2001model}
Mark~H Hansen and Bin~Yu and. 2001.
\newblock \href {https://doi.org/10.1198/016214501753168398} {Model selection and the principle of minimum description length}.
\newblock \emph{Journal of the American Statistical Association}, 96(454):746--774.

\bibitem[{He et~al.(2024)He, Doshi, Das, and Gromov}]{he2024learning}
Tianyu He, Darshil Doshi, Aritra Das, and Andrey Gromov. 2024.
\newblock \href {https://openreview.net/forum?id=aVh9KRZdRk} {Learning to grok: Emergence of in-context learning and skill composition in modular arithmetic tasks}.
\newblock In \emph{The Thirty-eighth Annual Conference on Neural Information Processing Systems}.

\bibitem[{Heimersheim and Nanda(2024)}]{heimersheim_how_2024}
Stefan Heimersheim and Neel Nanda. 2024.
\newblock How to use and interpret activation patching.
\newblock \emph{arXiv preprint arXiv:2404.15255}.

\bibitem[{Humayun et~al.(2024)Humayun, Balestriero, and Baraniuk}]{humayun2024grokking}
Ahmed~Imtiaz Humayun, Randall Balestriero, and Richard Baraniuk. 2024.
\newblock \href {https://openreview.net/forum?id=R2sVqqTf9p} {Grokking and the geometry of circuit formation}.
\newblock In \emph{ICML 2024 Workshop on Mechanistic Interpretability}.

\bibitem[{Jimenez et~al.(2024)Jimenez, Yang, Wettig, Yao, Pei, Press, and Narasimhan}]{jimenezswe}
Carlos~E Jimenez, John Yang, Alexander Wettig, Shunyu Yao, Kexin Pei, Ofir Press, and Karthik~R Narasimhan. 2024.
\newblock Swe-bench: Can language models resolve real-world github issues?
\newblock In \emph{The Twelfth International Conference on Learning Representations}.

\bibitem[{Kamradt(2023)}]{kamradt2023needle}
Greg Kamradt. 2023.
\newblock \href {https://github.com/gkamradt/LLMTest_NeedleInAHaystack} {{LLM Test - Needle in a Haystack}}.

\bibitem[{Kram{\'a}r et~al.(2024)Kram{\'a}r, Lieberum, Shah, and Nanda}]{kramar2024atp}
J{\'a}nos Kram{\'a}r, Tom Lieberum, Rohin Shah, and Neel Nanda. 2024.
\newblock Atp*: An efficient and scalable method for localizing llm behaviour to components.
\newblock \emph{arXiv preprint arXiv:2403.00745}.

\bibitem[{Kudo et~al.(2023)Kudo, Aoki, Kuribayashi, Brassard, Yoshikawa, Sakaguchi, and Inui}]{kudo-etal-2023-deep}
Keito Kudo, Yoichi Aoki, Tatsuki Kuribayashi, Ana Brassard, Masashi Yoshikawa, Keisuke Sakaguchi, and Kentaro Inui. 2023.
\newblock \href {https://doi.org/10.18653/v1/2023.eacl-main.98} {Do deep neural networks capture compositionality in arithmetic reasoning?}
\newblock In \emph{Proceedings of the 17th Conference of the European Chapter of the Association for Computational Linguistics}, pages 1351--1362, Dubrovnik, Croatia. Association for Computational Linguistics.

\bibitem[{Lee et~al.(2024)Lee, Bai, Pres, Wattenberg, Kummerfeld, and Mihalcea}]{lee2024a}
Andrew Lee, Xiaoyan Bai, Itamar Pres, Martin Wattenberg, Jonathan~K. Kummerfeld, and Rada Mihalcea. 2024.
\newblock \href {https://openreview.net/forum?id=dBqHGZPGZI} {A mechanistic understanding of alignment algorithms: A case study on {DPO} and toxicity}.
\newblock In \emph{Forty-first International Conference on Machine Learning}.

\bibitem[{Li et~al.(2023)Li, Bubeck, Eldan, Giorno, Gunasekar, and Lee}]{li2023textbooksneediiphi15}
Yuanzhi Li, Sébastien Bubeck, Ronen Eldan, Allie~Del Giorno, Suriya Gunasekar, and Yin~Tat Lee. 2023.
\newblock \href {https://arxiv.org/abs/2309.05463} {Textbooks are all you need ii: phi-1.5 technical report}.
\newblock \emph{Preprint}, arXiv:2309.05463.

\bibitem[{Lieberum et~al.(2023)Lieberum, Rahtz, Kram{\'a}r, Nanda, Irving, Shah, and Mikulik}]{lieberum2023does}
Tom Lieberum, Matthew Rahtz, J{\'a}nos Kram{\'a}r, Neel Nanda, Geoffrey Irving, Rohin Shah, and Vladimir Mikulik. 2023.
\newblock Does circuit analysis interpretability scale? evidence from multiple choice capabilities in chinchilla.
\newblock \emph{arXiv preprint arXiv:2307.09458}.

\bibitem[{Maaten and Hinton(2008)}]{maaten_visualizing_2008}
Laurens van~der Maaten and Geoffrey Hinton. 2008.
\newblock \href {http://jmlr.org/papers/v9/vandermaaten08a.html} {Visualizing {Data} using t-{SNE}}.
\newblock \emph{Journal of Machine Learning Research}, 9(86):2579--2605.

\bibitem[{Meng et~al.(2022)Meng, Bau, Andonian, and Belinkov}]{meng_locating_2022}
Kevin Meng, David Bau, Alex~J. Andonian, and Yonatan Belinkov. 2022.
\newblock \href {https://openreview.net/forum?id=-h6WAS6eE4} {Locating and {Editing} {Factual} {Associations} in {GPT}}.
\newblock In \emph{Advances in {Neural} {Information} {Processing} {Systems}}.

\bibitem[{Merullo et~al.(2023)Merullo, Eickhoff, and Pavlick}]{merullo_circuit_2023}
Jack Merullo, Carsten Eickhoff, and Ellie Pavlick. 2023.
\newblock \href {https://openreview.net/forum?id=fpoAYV6Wsk} {Circuit {Component} {Reuse} {Across} {Tasks} in {Transformer} {Language} {Models}}.
\newblock In \emph{The {Twelfth} {International} {Conference} on {Learning} {Representations}}.

\bibitem[{Miller et~al.(2024)Miller, Chughtai, and Saunders}]{miller2024transformer}
Joseph Miller, Bilal Chughtai, and William Saunders. 2024.
\newblock \href {https://openreview.net/forum?id=zSf8PJyQb2} {Transformer circuit evaluation metrics are not robust}.
\newblock In \emph{First Conference on Language Modeling}.

\bibitem[{Nanda(2023)}]{nanda_attribution_2023}
Neel Nanda. 2023.
\newblock Attribution patching: {Activation} patching at industrial scale.
\newblock \emph{URL: https://www. neelnanda. io/mechanistic-interpretability/attribution-patching}.

\bibitem[{Nanda et~al.(2022)Nanda, Chan, Lieberum, Smith, and Steinhardt}]{nanda_progress_2022}
Neel Nanda, Lawrence Chan, Tom Lieberum, Jess Smith, and Jacob Steinhardt. 2022.
\newblock \href {https://openreview.net/forum?id=9XFSbDPmdW} {Progress measures for grokking via mechanistic interpretability}.
\newblock In \emph{The {Eleventh} {International} {Conference} on {Learning} {Representations}}.

\bibitem[{Nikankin et~al.(2025)Nikankin, Reusch, Mueller, and Belinkov}]{nikankin2025arithmetic}
Yaniv Nikankin, Anja Reusch, Aaron Mueller, and Yonatan Belinkov. 2025.
\newblock \href {https://openreview.net/forum?id=O9YTt26r2P} {Arithmetic without algorithms: Language models solve math with a bag of heuristics}.
\newblock In \emph{The Thirteenth International Conference on Learning Representations}.

\bibitem[{Olah et~al.(2020)Olah, Cammarata, Schubert, Goh, Petrov, and Carter}]{olah2020zoom}
Chris Olah, Nick Cammarata, Ludwig Schubert, Gabriel Goh, Michael Petrov, and Shan Carter. 2020.
\newblock \href {https://doi.org/10.23915/distill.00024.001} {Zoom in: An introduction to circuits}.
\newblock \emph{Distill}.
\newblock Https://distill.pub/2020/circuits/zoom-in.

\bibitem[{Olsson et~al.(2022)Olsson, Elhage, Nanda, Joseph, DasSarma, Henighan, Mann, Askell, Bai, Chen, Conerly, Drain, Ganguli, Hatfield-Dodds, Hernandez, Johnston, Jones, Kernion, Lovitt, Ndousse, Amodei, Brown, Clark, Kaplan, McCandlish, and Olah}]{olsson2022context}
Catherine Olsson, Nelson Elhage, Neel Nanda, Nicholas Joseph, Nova DasSarma, Tom Henighan, Ben Mann, Amanda Askell, Yuntao Bai, Anna Chen, Tom Conerly, Dawn Drain, Deep Ganguli, Zac Hatfield-Dodds, Danny Hernandez, Scott Johnston, Andy Jones, Jackson Kernion, Liane Lovitt, Kamal Ndousse, Dario Amodei, Tom Brown, Jack Clark, Jared Kaplan, Sam McCandlish, and Chris Olah. 2022.
\newblock In-context learning and induction heads.
\newblock \emph{Transformer Circuits Thread}.
\newblock Https://transformer-circuits.pub/2022/in-context-learning-and-induction-heads/index.html.

\bibitem[{Otsuka and Saigo(2022)}]{otsuka_equivalence_2022}
Jun Otsuka and Hayato Saigo. 2022.
\newblock \href {https://proceedings.mlr.press/v177/otsuka22a.html} {On the {Equivalence} of {Causal} {Models}: {A} {Category}-{Theoretic} {Approach}}.
\newblock In \emph{Proceedings of the {First} {Conference} on {Causal} {Learning} and {Reasoning}}, volume 177 of \emph{Proceedings of {Machine} {Learning} {Research}}, pages 634--646. PMLR.

\bibitem[{Pearl(2009)}]{pearl2009causality}
Judea Pearl. 2009.
\newblock \emph{Causality}.
\newblock Cambridge University Press.

\bibitem[{Power et~al.(2022)Power, Burda, Edwards, Babuschkin, and Misra}]{power_grokking_2022}
Alethea Power, Yuri Burda, Harri Edwards, Igor Babuschkin, and Vedant Misra. 2022.
\newblock \href {https://arxiv.org/abs/2201.02177} {Grokking: {Generalization} {Beyond} {Overfitting} on {Small} {Datasets}}.

\bibitem[{Rivière et~al.(2024)Rivière, Pathak, Sessa, Hardin, Bhupatiraju, Hussenot, Mesnard, Shahriari et~al.}]{gemma}
Morgane Rivière, Shreya Pathak, Pier~Giuseppe Sessa, Cassidy Hardin, Surya Bhupatiraju, Léonard Hussenot, Thomas Mesnard, Bobak Shahriari, et~al. 2024.
\newblock \href {https://doi.org/10.48550/arXiv.2408.00118} {Gemma 2: Improving open language models at a practical size}.
\newblock \emph{CoRR}, abs/2408.00118.

\bibitem[{Sefidgaran et~al.(2023)Sefidgaran, Zaidi, and Krasnowski}]{sefidgaran_minimum_2023}
Milad Sefidgaran, Abdellatif Zaidi, and Piotr Krasnowski. 2023.
\newblock \href {https://proceedings.neurips.cc/paper_files/paper/2023/file/054e9f9a286671ababa3213d6e59c1c2-Paper-Conference.pdf} {Minimum {Description} {Length} and {Generalization} {Guarantees} for {Representation} {Learning}}.
\newblock In \emph{Advances in {Neural} {Information} {Processing} {Systems}}, volume~36, pages 1489--1525. Curran Associates, Inc.

\bibitem[{Shalev-Shwartz and Ben-David(2014)}]{shalev-shwartz_understanding_2014}
Shai Shalev-Shwartz and Shai Ben-David. 2014.
\newblock \emph{Understanding machine learning: {From} theory to algorithms}.
\newblock Cambridge university press.

\bibitem[{Shi et~al.(2024)Shi, Beltran-Velez, Nazaret, Zheng, Garriga-Alonso, Jesson, Makar, and Blei}]{shi2024hypothesis}
Claudia Shi, Nicolas Beltran-Velez, Achille Nazaret, Carolina Zheng, Adri\`{a} Garriga-Alonso, Andrew Jesson, Maggie Makar, and David~M. Blei. 2024.
\newblock \href {https://proceedings.neurips.cc/paper_files/paper/2024/file/abccb8a90b30d45b948360ba41f5a20f-Paper-Conference.pdf} {Hypothesis testing the circuit hypothesis in llms}.
\newblock In \emph{Advances in Neural Information Processing Systems}, volume~37, pages 94539--94567. Curran Associates, Inc.

\bibitem[{Srivastava et~al.(2023)Srivastava, Rastogi, Rao, Shoeb, Abid, Fisch, Brown et~al.}]{srivastava2023beyond}
Aarohi Srivastava, Abhinav Rastogi, Abhishek Rao, Abu Awal~Md Shoeb, Abubakar Abid, Adam Fisch, Adam~R. Brown, et~al. 2023.
\newblock \href {https://openreview.net/forum?id=uyTL5Bvosj} {Beyond the imitation game: Quantifying and extrapolating the capabilities of language models}.
\newblock \emph{Transactions on Machine Learning Research}.
\newblock Featured Certification.

\bibitem[{Stolfo et~al.(2023)Stolfo, Belinkov, and Sachan}]{stolfo-etal-2023-mechanistic}
Alessandro Stolfo, Yonatan Belinkov, and Mrinmaya Sachan. 2023.
\newblock \href {https://doi.org/10.18653/v1/2023.emnlp-main.435} {A mechanistic interpretation of arithmetic reasoning in language models using causal mediation analysis}.
\newblock In \emph{Proceedings of the 2023 Conference on Empirical Methods in Natural Language Processing}, pages 7035--7052, Singapore. Association for Computational Linguistics.

\bibitem[{Sun et~al.(2024)Sun, Ma, Ge, and Vosoughi}]{sun2024achieving}
Alan Sun, Chiyu Ma, Kenneth Ge, and Soroush Vosoughi. 2024.
\newblock \href {https://openreview.net/forum?id=v07KRLYxDX} {Achieving domain-independent certified robustness via knowledge continuity}.
\newblock In \emph{The Thirty-eighth Annual Conference on Neural Information Processing Systems}.

\bibitem[{Suzgun et~al.(2023)Suzgun, Scales, Sch{\"a}rli, Gehrmann, Tay, Chung, Chowdhery, Le, Chi, Zhou, and Wei}]{suzgun-etal-2023-challenging}
Mirac Suzgun, Nathan Scales, Nathanael Sch{\"a}rli, Sebastian Gehrmann, Yi~Tay, Hyung~Won Chung, Aakanksha Chowdhery, Quoc Le, Ed~Chi, Denny Zhou, and Jason Wei. 2023.
\newblock \href {https://doi.org/10.18653/v1/2023.findings-acl.824} {Challenging {BIG}-bench tasks and whether chain-of-thought can solve them}.
\newblock In \emph{Findings of the Association for Computational Linguistics: ACL 2023}, pages 13003--13051, Toronto, Canada. Association for Computational Linguistics.

\bibitem[{Syed et~al.(2024)Syed, Rager, and Conmy}]{syed2023attribution}
Aaquib Syed, Can Rager, and Arthur Conmy. 2024.
\newblock \href {https://doi.org/10.18653/v1/2024.blackboxnlp-1.25} {Attribution patching outperforms automated circuit discovery}.
\newblock In \emph{Proceedings of the 7th BlackboxNLP Workshop: Analyzing and Interpreting Neural Networks for NLP}, pages 407--416, Miami, Florida, US. Association for Computational Linguistics.

\bibitem[{Tigges et~al.(2024)Tigges, Hanna, Yu, and Biderman}]{tigges_llm_2024}
Curt Tigges, Michael Hanna, Qinan Yu, and Stella Biderman. 2024.
\newblock \href {https://openreview.net/forum?id=3Ds5vNudIE} {{LLM} {Circuit} {Analyses} {Are} {Consistent} {Across} {Training} and {Scale}}.
\newblock In \emph{The {Thirty}-eighth {Annual} {Conference} on {Neural} {Information} {Processing} {Systems}}.

\bibitem[{Vaswani et~al.(2017)Vaswani, Shazeer, Parmar, Uszkoreit, Jones, Gomez, Kaiser, and Polosukhin}]{vaswani_attention_2017}
Ashish Vaswani, Noam Shazeer, Niki Parmar, Jakob Uszkoreit, Llion Jones, Aidan~N Gomez, Łukasz Kaiser, and Illia Polosukhin. 2017.
\newblock Attention is all you need.
\newblock In \emph{Advances in {Neural} {Information} {Processing} {Systems}}, volume~30.

\bibitem[{Vig et~al.(2020)Vig, Gehrmann, Belinkov, Qian, Nevo, Singer, and Shieber}]{vig_investigating_2020}
Jesse Vig, Sebastian Gehrmann, Yonatan Belinkov, Sharon Qian, Daniel Nevo, Yaron Singer, and Stuart Shieber. 2020.
\newblock \href {https://proceedings.neurips.cc/paper/2020/file/92650b2e92217715fe312e6fa7b90d82-Paper.pdf} {Investigating gender bias in language models using causal mediation analysis}.
\newblock In \emph{Advances in neural information processing systems}, volume~33, pages 12388--12401.

\bibitem[{Vilas et~al.(2024)Vilas, Adolfi, Poeppel, and Roig}]{vilas_position_2024}
Martina~G. Vilas, Federuci Adolfi, David Poeppel, and Gemma Roig. 2024.
\newblock \href {https://arxiv.org/pdf/2406.01352} {Position: {An} {Inner} {Interpretability} {Framework} for {AI} {Inspired} by {Lessons} from {Cognitive} {Science}}.
\newblock In \emph{Proceedings of the 41st {International} {Conference} on {Machine} {Learning}}.

\bibitem[{Wang et~al.(2022)Wang, Variengien, Conmy, Shlegeris, and Steinhardt}]{wang_interpretability_2022}
Kevin~Ro Wang, Alexandre Variengien, Arthur Conmy, Buck Shlegeris, and Jacob Steinhardt. 2022.
\newblock \href {https://openreview.net/forum?id=NpsVSN6o4ul} {Interpretability in the {Wild}: a {Circuit} for {Indirect} {Object} {Identification} in {GPT}-2 {Small}}.
\newblock In \emph{The {Eleventh} {International} {Conference} on {Learning} {Representations}}.

\bibitem[{Wei et~al.(2022)Wei, Wang, Schuurmans, Bosma, ichter, Xia, Chi, Le, and Zhou}]{wei_chain--thought_2022}
Jason Wei, Xuezhi Wang, Dale Schuurmans, Maarten Bosma, brian ichter, Fei Xia, Ed~Chi, Quoc~V Le, and Denny Zhou. 2022.
\newblock \href {https://proceedings.neurips.cc/paper_files/paper/2022/file/9d5609613524ecf4f15af0f7b31abca4-Paper-Conference.pdf} {Chain-of-{Thought} {Prompting} {Elicits} {Reasoning} in {Large} {Language} {Models}}.
\newblock In \emph{Advances in {Neural} {Information} {Processing} {Systems}}, volume~35, pages 24824--24837. Curran Associates, Inc.

\bibitem[{Wiedemer et~al.(2023)Wiedemer, Mayilvahanan, Bethge, and Brendel}]{wiedemer2023compositional}
Thadd{\"a}us Wiedemer, Prasanna Mayilvahanan, Matthias Bethge, and Wieland Brendel. 2023.
\newblock \href {https://openreview.net/forum?id=LqOQ1uJmSx} {Compositional generalization from first principles}.
\newblock In \emph{Thirty-seventh Conference on Neural Information Processing Systems}.

\bibitem[{Wu et~al.(2023)Wu, Geiger, Icard, Potts, and Goodman}]{wu2023interpretability}
Zhengxuan Wu, Atticus Geiger, Thomas Icard, Christopher Potts, and Noah Goodman. 2023.
\newblock \href {https://proceedings.neurips.cc/paper_files/paper/2023/file/f6a8b109d4d4fd64c75e94aaf85d9697-Paper-Conference.pdf} {Interpretability at scale: Identifying causal mechanisms in alpaca}.
\newblock In \emph{Advances in Neural Information Processing Systems}, volume~36, pages 78205--78226. Curran Associates, Inc.

\bibitem[{Ye et~al.(2021)Ye, Xie, Cai, Li, Li, and Wang}]{ye2021towards}
Haotian Ye, Chuanlong Xie, Tianle Cai, Ruichen Li, Zhenguo Li, and Liwei Wang. 2021.
\newblock \href {https://proceedings.neurips.cc/paper_files/paper/2021/file/c5c1cb0bebd56ae38817b251ad72bedb-Paper.pdf} {Towards a theoretical framework of out-of-distribution generalization}.
\newblock In \emph{Advances in Neural Information Processing Systems}, volume~34, pages 23519--23531. Curran Associates, Inc.

\bibitem[{Yu et~al.(2024)Yu, Kaur, Gupta, Brown-Cohen, Goyal, and Arora}]{yu2024skillmix}
Dingli Yu, Simran Kaur, Arushi Gupta, Jonah Brown-Cohen, Anirudh Goyal, and Sanjeev Arora. 2024.
\newblock \href {https://openreview.net/forum?id=Jf5gplvglq} {{SKILL}-{MIX}: a flexible and expandable family of evaluations for {AI} models}.
\newblock In \emph{The Twelfth International Conference on Learning Representations}.

\bibitem[{Zhong et~al.(2023)Zhong, Liu, Tegmark, and Andreas}]{zhong_clock_2023}
Zhiqian Zhong, Ziming Liu, Max Tegmark, and Jacob Andreas. 2023.
\newblock \href {https://arxiv.org/pdf/2306.17844} {The {Clock} and the {Pizza}: {Two} {Stories} in {Mechanistic} {Explanation} of {Neural} {Networks}}.
\newblock In \emph{Advances in {Neural} {Information} {Processing} {Systems}}.

\end{thebibliography}
